\documentclass[preprint,12pt,a4paper,number,nopreprintline]{elsarticle}
\usepackage[T1]{fontenc}
\usepackage{lmodern}
\input{glyphtounicode}\pdfgentounicode=1
\usepackage{amsmath,amssymb,booktabs,array,longtable,graphicx,setspace,etoolbox}
\usepackage[a4paper,left=22mm,right=22mm,top=25mm,bottom=25mm,includefoot]{geometry}
\usepackage{float,placeins}
\usepackage[font=footnotesize,skip=5pt,singlelinecheck=false,hypcap=false]{caption}
\usepackage[hidelinks]{hyperref}
\makeatletter
\let\ps@pprintTitle\ps@plain
\renewcommand\elsparagraph{\@startsection{paragraph}{4}{0\z@}%
  {10\p@ \@plus 6\p@ \@minus 3\p@}{-6\p@}%
  {\normalfont\normalsize\bfseries}}
\patchcmd{\pprintMaketitle}{\elsaddress\par\vskip36pt}{\elsaddress\par\vskip4pt}{}{\PackageError{author-layout}{Frontmatter spacing patch failed}{}}
\patchcmd{\pprintMaketitle}{\@title\par\vskip18pt}{\@title\par\vskip10pt}{}{\PackageError{author-layout}{Title spacing patch failed}{}}
\makeatother
\DeclareCaptionFont{reviewspacing}{\doublespacing}
\AtBeginEnvironment{table}{\singlespacing}
\AtBeginEnvironment{longtable}{\singlespacing}

\newcommand{\clip}{\operatorname{clip}}

\newcommand{\LL}{\mathrm{LL}}
\newcommand{\FAIL}{\mathrm{fail}}

\newcommand{\ALL}{E_{\mathrm{all}}}
\newcommand{\IN}{E_{\mathrm{in}}}

\begin{document}
\singlespacing
\begin{frontmatter}

\title{From Failure Prediction to Component Selection in Semantic Segmentation}
\renewcommand{\thefnote}{$\dagger$}
\author[jx]{Minhui Kang\fnref{equal}\corref{cor1}}
\author[jx]{Yiyao Zhang\fnref{equal}}
\author[sd]{Peizhe Long}
\author[jx]{Jiafeng Qi}
\author[nchu]{Hongyu Jin}
\author[jx]{Jiayun Zheng}
\address[jx]{School of Power Engineering, Jiangxi Electric Vocational and Technical College, Nanchang, Jiangxi, China}
\address[sd]{College of Information Science and Engineering, Shandong Agricultural University, Tai'an, Shandong, China}
\address[nchu]{College of Aviation Manufacturing Industry, Nanchang Hangkong University, Jingdezhen, Jiangxi, China}
\fntext[equal]{These authors contributed equally to this work.}
\cortext[cor1]{Corresponding author: \href{mailto:zebukang@hotmail.com}{zebukang@hotmail.com}.}
\hypersetup{pdfauthor={Minhui Kang, Yiyao Zhang, Peizhe Long, Jiafeng Qi, Hongyu Jin, Jiayun Zheng}}

\begin{abstract}
\doublespacing
\typeout{PR-ABSTRACT-BASELINE=\the\baselineskip}
Reliable scene understanding requires deciding which predictions to trust, retain, or inspect. In semantic segmentation, component failure predictors support these decisions, but their predictive performance does not fully describe the value of the resulting selection. We develop a decision-centered evaluation framework that connects three questions: what an available score predicts, what can be learned from its existing fields, and what its ranking delivers. Across three segmentation systems on Cityscapes and ACDC at 90\% retention, adapted MetaSeg reduces retained failure relative to forecast entropy in all twelve architecture--population comparisons, but does not establish a failure advantage over direct probability ranking. Image-nested source experiments reveal a different opportunity: reusing the full field set improves predictive and retained-loss performance under both learners in all three systems, while the raw OCRNet probe improves prediction with unresolved selective gains. Mapping a fixed prediction pair to conditional log-loss risk produces mixed selection effects, which we examine through the components exchanged by the two rankings. These findings distinguish the performance of a fitted score from the value obtainable from its inputs and the consequences of its use. The framework provides a practical basis for deciding whether to adopt an additional score, relearn existing fields, or revise a selection rule, and for explaining the gains and failures of component-level reliability methods.

\end{abstract}
\end{frontmatter}

\doublespacing\normalsize
\typeout{PR-TEXTWIDTH=\the\textwidth; PR-BODY-BASELINE=\the\baselineskip}
\section{Introduction}
Semantic segmentation supports scene understanding in automated driving, where a system must decide which predicted regions to trust, retain or inspect. We treat each eight-connected region of a single class in the frozen segmentation map as a component, and use component failure predictors to estimate errors on these regions~\cite{rottmann2020}. Their scores can prioritize inspection or select regions under a fixed component budget. For these decisions, the value of an additional score lies in the components it retains compared with an alternative already available to the system. Predictive performance provides evidence about the score; evaluating the resulting selection establishes what using it delivers.

Our results show why this distinction matters. In DeepLab/P1, adapted MetaSeg has a higher pooled AUROC than the base failure predictor, 0.8420 versus 0.8375, yet retains a higher image-averaged failure rate at 90\% coverage, 24.709\% versus 24.464\%. Against entropy of the same base forecast, it instead reduces retained failure by 1.655 percentage points. Thus the same score receives different selective assessments depending on the reference. This example compares pooled discrimination with image-macro selection on all-valid components; the metrics differ in both functional and weighting. Paired uncertainty and eligibility sensitivity are reported in Section~\ref{sec:results}.

\begin{figure}[!t]\centering
\includegraphics[width=\linewidth]{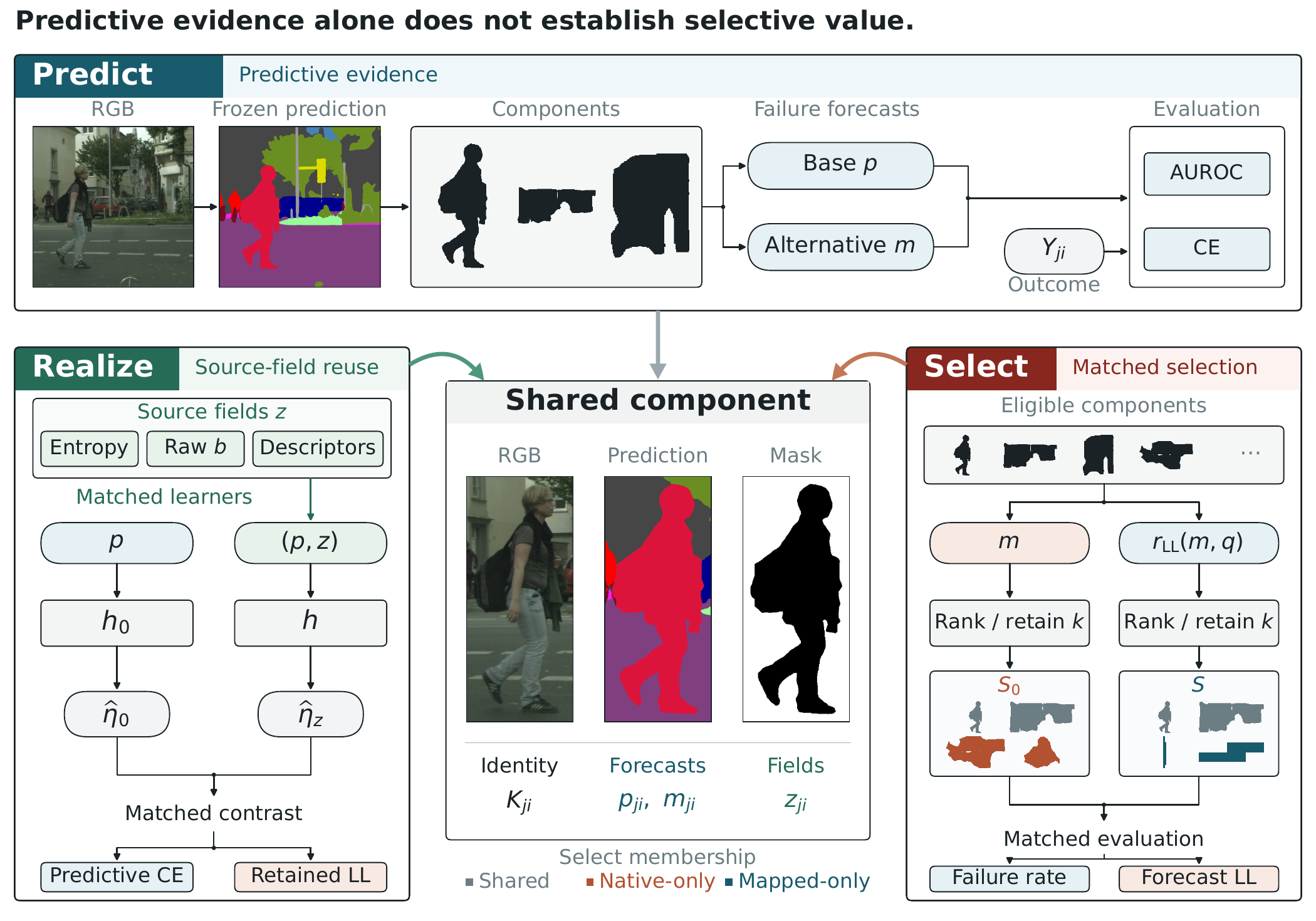}
\caption{\textbf{Three evidence questions for component reliability.} Predict evaluates forecasts; Realize tests added field information; Select compares retained sets under the same budget and loss. Arrows connect the shared component record. The Select example compares native $m$ with mapped $r_{\LL}(m,q)$, evaluating fixed $p$. Gray/orange/blue masks denote shared/native-only/mapped-only members; these unordered subsets use independent display scales.}
\label{fig:questions}\end{figure}

The distinction also guides how to improve a reliability method. A weak fitted score can leave useful information in its inputs, while a better probability forecast can still have unresolved selection benefits. These possibilities call for different responses: revising the score, relearning existing information, or changing how predictions are used to select components. We organize their evaluation around three questions (Fig.~\ref{fig:questions}):
\begin{itemize}
\item \textbf{Predict:} How well does an available score forecast component failure?
\item \textbf{Realize:} What additional predictive and selective value can be learned from its existing fields?
\item \textbf{Select:} Does its ranking improve the intended retained-set loss over an available reference at the same budget?
\end{itemize}

In Figure~\ref{fig:questions}, the black silhouettes show individual component masks extracted from the full prediction. They are displayed separately at independent scales; the RGB and prediction thumbnails show only a crop of that scene.

The common component record connects these questions. Predict evaluates its forecasts, Realize tests field reuse with matched probability-only controls, and Select compares the sets retained by alternative rankings. The right-hand panel of Figure~\ref{fig:questions} illustrates Select through native-versus-risk-mapped ranking; the broader question also covers comparing an additional score with an existing probability or entropy ranking. In each case, the comparison asks whether the changed selection improves the intended objective.

\begin{samepage}
We distinguish failure screening from forecast auditing. The first measures failures among retained segmentation components; the second evaluates their recorded failure probabilities. A correctly anticipated failure can have low forecast loss while remaining an undesirable segmentation result. Holding the evaluated probabilities fixed lets us attribute changes in forecast log loss to the selected components. The two endpoints therefore describe different aspects of selection value.\par
\end{samepage}

Conditional-risk theory characterizes optimal selection~\cite{franc2023}. Research on uncertainty evaluation emphasizes the intended downstream use~\cite{kahl2024,zenk2025,traub2024}. Building on these foundations, we study three fixed segmentation systems on Cityscapes and ACDC and make three contributions:

\paragraph{Reference-aware comparisons}
MetaSeg reduces retained failure relative to forecast entropy in all twelve architecture--population comparisons, while increasing retained forecast loss; it establishes no failure advantage over direct probability ranking. These matched comparisons show why the available reference and the intended loss are essential to deciding whether an additional score is useful.

\paragraph{Controlled feature reuse}
Image-nested source experiments separate the current score from the value obtainable from its fields. Reusing the full field set improves predictive and retained-loss performance under both specified learners in all three systems, while the OCRNet probe adds predictive information with unresolved selective gains. The contrast identifies an opportunity to improve reliability by reusing existing information.

\paragraph{Explaining selection changes}
With fixed forecasts, risk mapping produces mixed primary effects despite its loss-aligned target. Coverage sensitivity and exact component-exchange accounting connect these effects to which components enter and leave the retained set and the losses they carry. This provides a concrete explanation of how a ranking change gains or loses value, beyond observing that it changes the selection.

\section{Related Work}
Reliability research provides ways to estimate uncertainty, predict failure and select outputs. Our study connects these lines of work through a practical question: how can their evidence guide whether to adopt a score, relearn its inputs or revise its use in selection?

\paragraph{Uncertainty estimation and segmentation reliability}
Predictive uncertainty can be estimated through Monte Carlo dropout or deep ensembles~\cite{gal2016,lakshminarayanan2017}, with aleatoric and epistemic uncertainty distinguished in dense vision tasks~\cite{kendall2017}. Distribution-shift benchmarks extend evaluation beyond the training distribution~\cite{ovadia2019}. The resulting assessments depend on the task: de Jorge et al. report differences across segmentation robustness, calibration, misclassification and out-of-distribution detection~\cite{dejorge2023}, while Zhu et al. show that calibration and out-of-distribution methods can impair classification failure detection~\cite{zhu2024}. These findings motivate examining what a reliability score is used to accomplish. We focus on its value for retaining segmentation components under a specified loss and per-image budget.

\paragraph{Failure prediction and confidence aggregation}
Moving from a pixel or sample score to a regional decision requires choosing both the information to use and how to aggregate it. Maximum softmax probability provides a standard detection baseline~\cite{hendrycks2017}, while learned estimators can target correctness-related quantities such as true-class probability~\cite{corbiere2019}. MetaSeg combines aggregated softmax dispersion and geometry to classify zero-overlap components and regress component quality~\cite{rottmann2020}. Medical-segmentation benchmarks examine aggregation for image-level failure detection under shift~\cite{zenk2025}; ValUES separates uncertainty-method components and evaluates their downstream uses~\cite{kahl2024}. Building on this distinction between information and its use, we examine both the ranking delivered by an adapted MetaSeg classifier and the source value obtainable by relearning existing fields.

\paragraph{Selective prediction and reject-option learning}
Selective classification relates coverage to the risk of accepted predictions~\cite{elyaniv2010,hendrickx2024}. Selection can be applied after training to meet a risk target~\cite{geifman2017}, or learned jointly with the predictor, as in SelectiveNet~\cite{geifman2019}. Conditional-risk theory characterizes optimal strategies~\cite{franc2023}; subsequent evaluation includes covariate and label shifts~\cite{liang2024} and improved confidence estimates from fixed classifier logits~\cite{cattelan2024}. These approaches establish ranking construction as a substantive part of reliability. Our comparisons examine its incremental value at the component level: a native score, a model relearned from existing fields, and a remapping of fixed forecasts represent distinct ways to change the selection.

\paragraph{Calibration and decision-aware forecasts}
Calibration concerns the probabilities attached to predictions. Temperature scaling addresses neural-classifier miscalibration~\cite{guo2017}; local temperature scaling and selective scaling address spatial variation, overconfidence and domain shift in segmentation~\cite{ding2021,wang2023}. Decision-aware calibration specifies the decisions for which forecasts should be useful~\cite{zhao2021}. Here, the forecasted event is component failure, defined by zero overlap with same-class ground truth. We examine both the predictive quality of fitted failure probabilities and the log loss of a fixed base forecast on selected components. This separates improving a forecast from selecting records on which that forecast performs better.

\paragraph{Evaluation measures and loss alignment}
Proper scoring rules assess probability forecasts~\cite{gneiting2007}, while rejection measures and visualization spaces account for accepted and rejected decisions~\cite{condessa2017,hanczar2019}. Work on selective prediction and calibration~\cite{ding2020}, comparisons across 523 classifiers~\cite{galil2023}, and requirements for multi-threshold evaluation~\cite{traub2024} clarify the importance of the evaluation target. Soft Dice Confidence develops an image-level score aligned with Dice loss~\cite{borges2026}. We examine loss alignment through conditional-risk mapping of a fixed prediction pair. Comparing retained failure and fixed-forecast log loss reveals the different consequences for the two objectives; exchanged component identities explain the observed loss difference.

\paragraph{Position of this study}
Together, these comparisons support a diagnosis of where improvement can come from. Native-score comparisons assess whether an additional score delivers value over an available reference. Image-nested controls test the opportunity in existing source fields. Fixed-pair comparisons isolate the effect of changing the ranking rule, and component-exchange accounting explains the resulting loss difference. A weak score can therefore coexist with useful inputs, and a changed ranking can have favorable or adverse consequences. Our contribution is to connect these distinctions to evidence that guides the next intervention in a component-reliability system. Supplement S9 positions the comparisons by evaluation unit and target.

\section{Evaluation Framework}\label{sec:evaluation}
To evaluate what a reliability score delivers in component selection, we first define the component-level outcomes and the losses of interest. We then compare rankings at matched retention budgets and explain how the evaluation loss determines the corresponding conditional-risk target. These definitions connect predictive assessments to the components retained and the losses they incur.
\subsection{Component Failure and Evaluation Losses}
Let $K_{ji}$ be the $i$th eight-connected component of one predicted class in the frozen segmentation map of image $j$. Its binary outcome is $Y_{ji}=1$ when its maximum IoU with any same-class ground-truth component is zero after excluding ignore-255 pixels. A component with no valid supervision is omitted from outcome evaluation. This endpoint detects complete same-class non-overlap; a poorly segmented region with some overlap is not a failure under this definition. Component identities and the numerical tie convention of the archived maps remain fixed.

Each component carries a base failure forecast $p_{ji}$. A score $s_{ji}$ ranks components for selection, with larger values rejected first. Candidate and reference scores may differ, but their retained log losses always evaluate the same base forecast. Write $q_{ji}=\clip(p_{ji},10^{-6},1-10^{-6})$. The two losses are
\begin{equation}\label{eq:loss}
\ell_{\FAIL,ji}=Y_{ji},\qquad
\ell_{\LL,ji}=-Y_{ji}\log q_{ji}-(1-Y_{ji})\log(1-q_{ji}).
\end{equation}
Failure loss measures the failed segmentation components left after screening. Log loss, in nats, measures forecast quality on the retained probability records. Holding $p$ fixed attributes a change in retained LL to selection rather than to substituting or recalibrating the forecast. Lower retained LL describes those selected records; it does not establish global calibration. A ranking score need not itself have probability semantics.

\begin{figure}[!htbp]
\centering
\includegraphics[width=\linewidth,trim=0 12.90bp 0 4.88bp,clip]{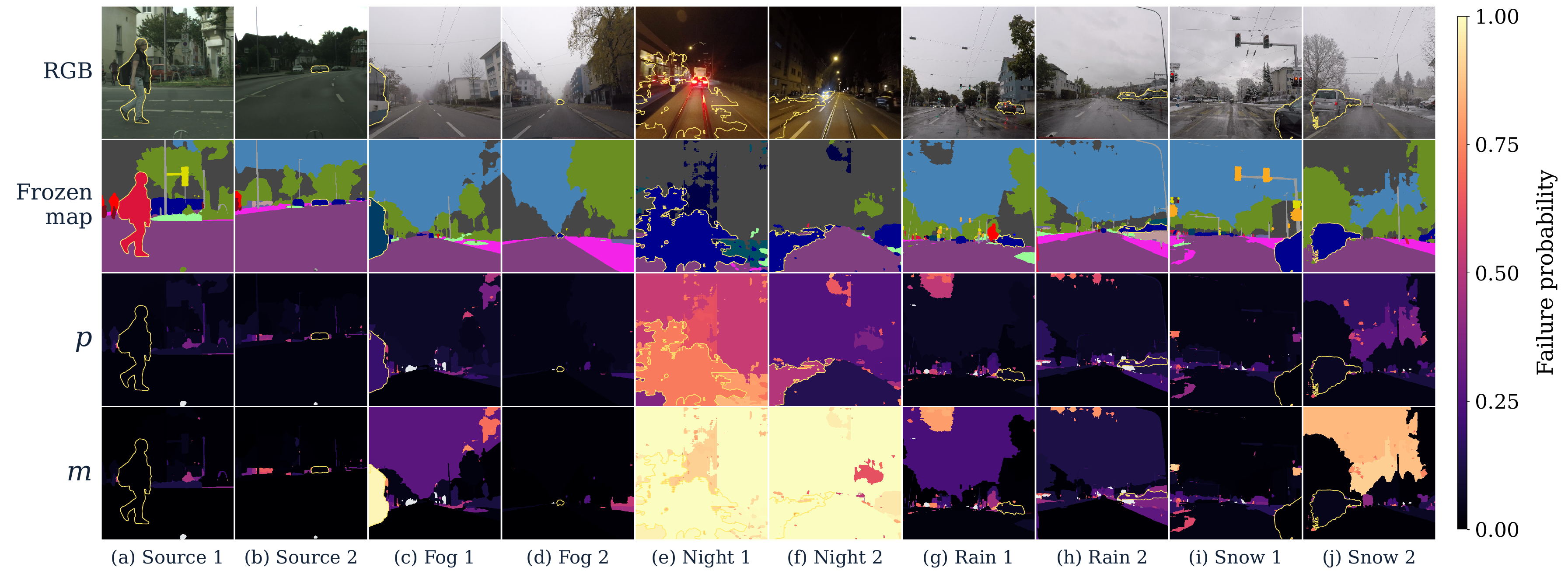}
\caption{Frozen DeepLab components and failure forecasts in ten Cityscapes/ACDC scenes. Rows show RGB, predicted classes, base failure probabilities $p$ and adapted MetaSeg probabilities $m$. Crops align within columns; probability maps share the $[0,1]$ scale. Gray marks excluded components under complete-image eligibility; yellow outlines mark predicted components.}\label{fig:components}
\end{figure}

Figure~\ref{fig:components} illustrates these objects in ten frozen street-scene predictions. It shows $p$ alongside the alternative failure probability $m$ supplied by the MetaSeg adaptation in Section~\ref{sec:study}. Both maps attach one probability to each predicted component, rather than a separate confidence to each pixel. The display contains two Cityscapes source-development images (P1) and two ACDC adverse-development images (P2) for each of fog, night, rain and snow. It retains the original scene in (a); additional scenes follow fixed image-identity ordering within population/weather strata, without selection on forecasts or outcomes.

\subsection{Matched Selection and Evaluation}
Matching retained counts within each image prevents a ranking from lowering loss simply by rejecting more components or reallocating rejection across images. For eligible components $E_j$, let $n_j=|E_j|$. At coverage $c$, every selector retains $k_j=\lceil cn_j\rceil$ components in image $j$. We use the executed descending-score, ascending-component-ID order, reject the first $n_j-k_j$ entries and retain the tail $S_j(s)$. This fixes boundary ties as well as cardinality. For nonempty eligible images $J_E$, define
\begin{equation}\label{eq:gain}
\begin{aligned}
L_\ell(s)&=\frac{1}{|J_E|}\sum_{j\in J_E}\frac{1}{k_j}
                 \sum_{i\in S_j(s)}\ell_{ji},\\
\Delta_\ell(s;s_0)&=L_\ell(s_0)-L_\ell(s).
\end{aligned}
\end{equation}
Positive gain favors the candidate $s$ over reference $s_0$. Images receive equal weight after within-image averaging. The comparison fixes the eligible component IDs, forecast, budget and loss within each population. Empty eligible images are recorded and omitted from the denominator. The primary coverage is $c=0.90$; the ceiling can make the realized retained fraction slightly larger.

\subsection{Conditional Risk and Ranking Rules}
Let $\eta_{ji}$ denote the conditional probability of failure given the information available for selection. Conditional expected losses for the two objectives are
\begin{equation}\label{eq:risk}
r_{\FAIL}(\eta)=\eta,\qquad
r_{\LL}(\eta,q)=-\eta\log q-(1-\eta)\log(1-q).
\end{equation}
At a fixed cardinality, retaining the smallest conditional risks minimizes conditional expected loss~\cite{franc2023}. This follows by exchanging a retained higher-risk component with a rejected lower-risk one; linearity of expectation is sufficient. When $\eta=q$, forecast-loss risk is entropy $H(q)$, whereas failure risk is $q$.

The empirical selectors use fitted probabilities rather than the unknown $\eta$. In particular, for an available failure predictor $m$ we compare its native ranking $s_{\rm native}=m$ with $s_{\rm map}=r_{\LL}(m,q)$ while holding $(p,m)$ fixed. The mapping combines the estimated failure probability $m$ with the base forecast $p$ whose loss is evaluated. Native-versus-mapped selection tests a change in ranking construction without refitting either forecast; comparison with entropy tests added value over a base-forecast rule. Its advantage under fitted probabilities is assessed by the retained-set comparison. The exact outcome formula, score conventions and a synthetic exchange illustrating the two objectives appear in Supplement S1.

\section{Method}\label{sec:study}
\subsection{Data and Segmentation Systems}
We study SegFormer-B0~\cite{xie2021}, DeepLabV3 with ResNetV1c-50-D8~\cite{chen2017}, and OCRNet with HRNet-W48~\cite{yuan2020}. Cityscapes~\cite{cordts2016} supplies 300 source-fitting images (P0) and 200 image-disjoint source-development images (P1). ACDC~\cite{sakaridis2021} supplies 406 adverse-development images (P2) and 400 final-population images (P3). Thus 1,306 unique images are shared across the three architectures. All segmentation maps and component identities are fixed.

\subsection{Failure Forecasts and Reference Scores}
We use two base-forecast references to distinguish failure screening from forecast auditing. Direct probability $p$ ranks likely failures; its entropy $H(p)$ targets forecast loss when $p$ equals the conditional failure probability (Eq.~\eqref{eq:risk}). Throughout these comparisons, ``entropy'' means the entropy of the component-failure forecast, not mean pixel entropy. The base predictor is an image-balanced L1 logistic model using eighteen size, geometry and probability descriptors and nineteen predicted-class indicators. The entropy reference retains its frozen source normalization and numerical convention. Supplement S2 specifies the preprocessing and fit.

The alternative forecast $m$ comes from a classification adaptation of MetaSeg~\cite{rottmann2020}, using fifteen dispersion/size descriptors and nineteen mean class probabilities. It fits sparse logistic classification on source components with nonempty interiors. Fits and inner-fold accuracy use equal component weights; image-grouped inner folds select regularization. Its eight-neighbor interior differs from the base predictor's four-neighbor boundary convention. Supplement S2.6 specifies these adaptation details, the five fixed regularization candidates and preprocessing.

Source predictions are out-of-fold, using five image folds. A single predictor fitted on the full source population supplies P1--P3. Learning from these predictions and other fitted fields requires the additional nesting described below.

The primary analysis evaluates all outcome-valid components ($\ALL$, 381,175 architecture--component instances). A sensitivity analysis restricts every comparator to nonempty interiors ($\IN$, 311,297 instances) and recomputes each image's retention budget. This checks whether the findings persist under the eligibility restriction used to train MetaSeg. Predictive AUROC pools components, whereas retained losses average within images and then give each image equal weight. Complete extraction and eligibility counts are in Supplement S4.

\subsection{Learning from Existing Fields}
\begin{figure}[!t]\centering
{\small\bfseries (a) Strict source controls\par}\smallskip
\includegraphics[width=\linewidth,trim=0 6.27bp 0 4.98bp,clip]{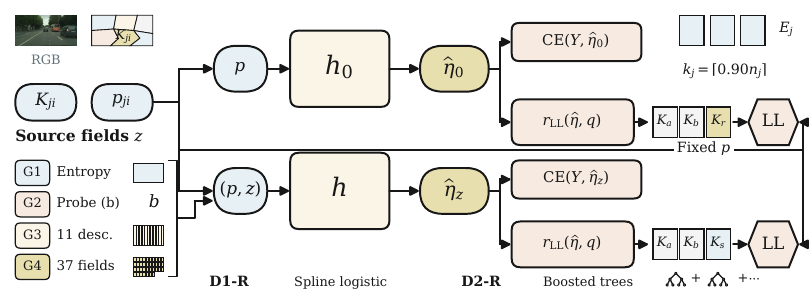}\par\medskip
\begin{minipage}[t]{.425\linewidth}\centering
{\small\bfseries (b) Raw probe\par}\smallskip
\includegraphics[width=\linewidth,trim=0 22.39bp 0 2.49bp,clip]{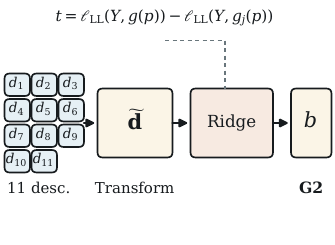}
\end{minipage}\hfill
\begin{minipage}[t]{.565\linewidth}\centering
{\small\bfseries (c) Image-nested construction\par}\smallskip
\includegraphics[width=\linewidth,trim=0 0.3bp 0 1.31bp,clip]{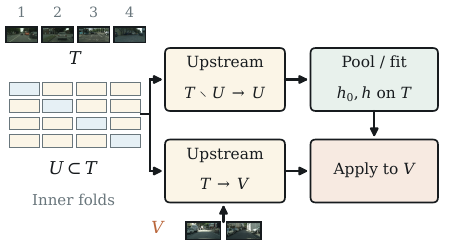}
\end{minipage}
\caption{Probe construction and matched source controls. (a) Matched learners produce diagnostic forecasts; CE evaluates prediction and mapped-risk ranking evaluates selection using fixed $p$. (b) Training calibration losses supply the ridge target; raw $b$ is a source field. (c) Image-nested inputs exclude $V$ labels from fitting. Thumbnails are archived P0 examples; other objects are schematic.}
\label{fig:source-method}
\end{figure}

A fitted score may leave useful information in its inputs. We test this possibility by comparing a learned $p$-only forecast, $\widehat\eta_0=h_0(p)$, with $\widehat\eta_z=h(p,z)$ within the same learner family (Fig.~\ref{fig:source-method}a). We assess two families separately: spline/interaction L2 logistic regression (D1-R) and histogram gradient boosting (D2-R).

The added fields are mean pixel entropy (G1), a raw calibration-opportunity probe (G2), the probe's eleven descriptors (G3), or all thirty-seven base-predictor fields (G4). These groups distinguish a learned scalar summary from its underlying descriptors and the full existing representation.

Two controls distinguish field reuse from a more flexible use of $p$. The learned $p$-only control (G0) allows the same learner to refit $p$. The dimension-matched $p$-derived control (PC) adds as many columns as $z$, using a fixed basis of $p$. PC matches column count, not the full function class. For G2, its added column is a copy of $p$; under D2-R, this gives the same contrast as G0. The basis and field-specific control names are given in Supplement S2.5.

Because some inputs are themselves learned, validation labels must be excluded from both input construction and the downstream learner (Fig.~\ref{fig:source-method}c). For each outer validation fold $V$, all fitting uses the other folds $T$. Inner-held-out upstream fits within $T$ generate the downstream training inputs; an upstream fit on all $T$ generates the validation inputs. Downstream transformations and learners are fitted only on $T$. Algorithm S1 gives the complete image-nested construction.

The raw probe $b$ summarizes a source-side calibration opportunity (Fig.~\ref{fig:source-method}b). A global affine-logit calibrator and an image-specific fit supply training losses; ridge regression predicts their signed difference from eleven existing geometry and probability descriptors. At application, only the fitted descriptor transform and ridge are used. The image-specific fit constructs the training target and is not an available selector. G2 supplies raw $b$ to the downstream learner; $b$ is neither a failure probability nor a direct retained-loss target. Supplement S2.3 gives its construction and its role in the original frozen score.

\FloatBarrier

Predictive gain measures the reduction in image-macro cross-entropy of $\widehat\eta_z$ relative to G0 and PC. Selective gain instead ranks components by $r_{\LL}(\widehat\eta_z,q)$ and evaluates retained loss against G0, PC and entropy rankings. The evaluated $p$ remains unchanged and is constructed without outer validation labels.

\subsection{Evaluation Protocol}
All rankings use the matched per-image counts and fixed-forecast losses defined in Section~\ref{sec:evaluation}, with 90\% retention as the primary setting. Reported intervals are pointwise 95\% paired-image bootstrap intervals from the registered 2,000 draws; candidate and reference share draws. They quantify evaluation-sample uncertainty conditional on the fitted predictors, excluding repeated-training variability. An interval crossing zero leaves the direction unresolved; it does not establish equivalence.\par

Sensitivity analyses restrict eligibility for MetaSeg and mapping, vary the mapping retention budget, and resample complete target sequences for mapping while preserving the image-weighted estimand. The sequence analysis checks dependence among target images; MetaSeg comparisons retain the paired-image analysis.

Only the original P3 frozen-score comparison matrix was specified before outcome reveal; subsequent P3 analyses, including MetaSeg and mapping, are retrospective. Supplementary Sections S7--S8 document additional analyses within this study, their shared data and the analysis chronology.

\section{Results}\label{sec:results}
\subsection{Selective comparisons depend on the reference and loss}
\begin{figure}[!t]\centering
\includegraphics[width=\linewidth]{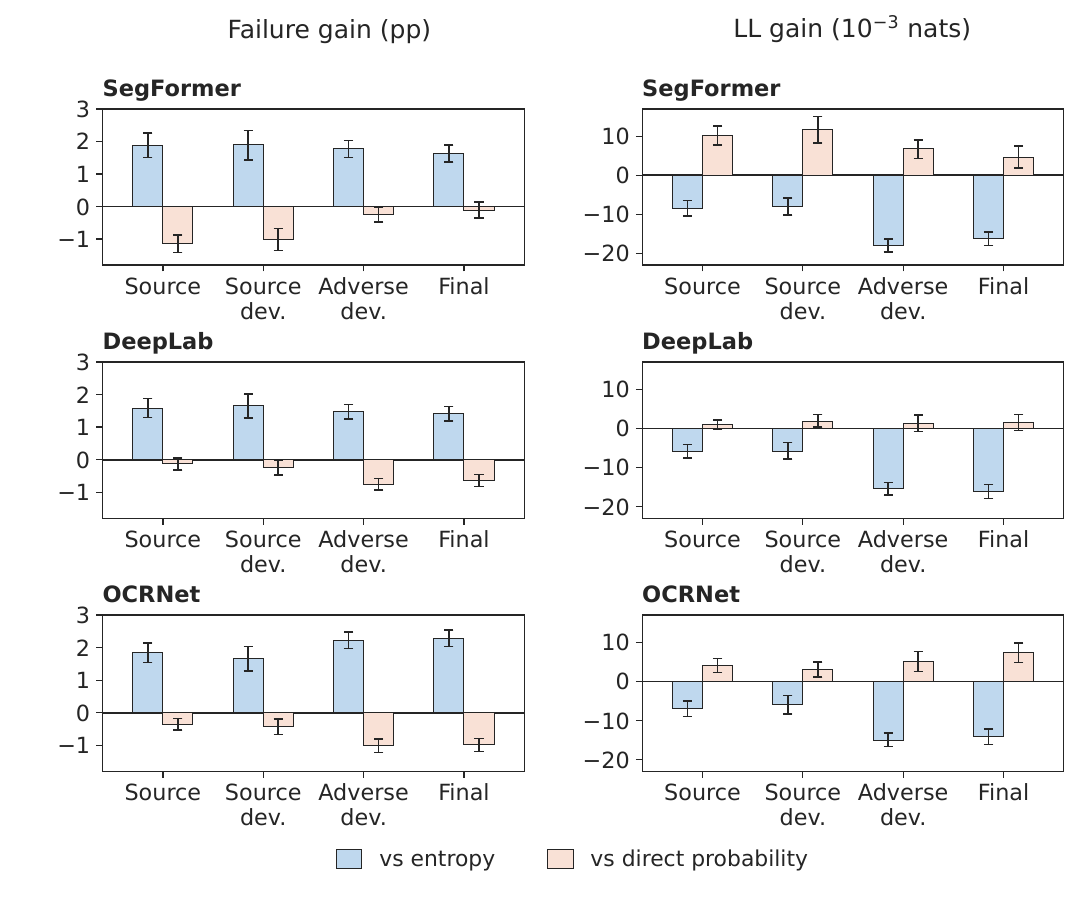}
\caption{MetaSeg gains against entropy (blue) and direct probability (peach) at 90\% all-valid retention. Rows denote architectures; columns show failure and forecast-loss gains. Positive favors MetaSeg; bars show pointwise 95\% paired-image intervals. Exact values: Supplement S4, Table S8.}\label{fig:comparators}
\end{figure}

The same MetaSeg ranking reduces retained failures relative to entropy but establishes no such advantage over direct probability (Fig.~\ref{fig:comparators}). Predictions, eligible components and retention budgets are unchanged across these comparisons. Against entropy, failure decreases in all twelve architecture--population cells, with favorable intervals, while forecast LL increases in every cell. Against direct $p$, ten failure intervals are adverse and two cross zero. The point estimates order $p$, MetaSeg and entropy from lower to higher failure, with the reverse order for forecast LL. Table~\ref{tab:primary-levels} reports absolute performance; Supplement S4, Table S8 gives the exact matched contrasts.

The reference therefore changes the conclusion about adding MetaSeg to a system that already has $p$. Beating entropy on failure screening does not establish added value over the available probability ranking. Forecast auditing gives a different assessment: against direct $p$, MetaSeg has nine favorable LL intervals and three crossing intervals. The intended loss and the available alternative are both needed to judge a score's selective value.

\FloatBarrier
\begin{table}[!htbp]
\centering
\begingroup\setstretch{1}\fontsize{9.5}{11.5}\selectfont
\setlength{\tabcolsep}{3pt}\renewcommand{\arraystretch}{1.00}
\textbf{(a) Failure prediction}\par\smallskip
\begin{tabular*}{\linewidth}{@{\extracolsep{\fill}}lrrrrrr@{}}\toprule
Population & Images & $N$ & AUROC $p\uparrow$ & AUROC $m\uparrow$ & CE $p\downarrow$ & CE $m\downarrow$\\\midrule
\multicolumn{7}{l}{\textit{SegFormer}}\\[1pt]
P0 & 300 & 18,728 & 0.8105 & 0.7157 & 0.4993 & 0.6076\\
P1 & 200 & 12,209 & 0.8115 & 0.7361 & 0.5121 & 0.5927\\
P2 & 406 & 32,429 & 0.7562 & 0.7090 & 0.5834 & 0.6825\\
P3 & 400 & 29,614 & 0.7403 & 0.6973 & 0.5953 & 0.6880\\\addlinespace[3pt]
\multicolumn{7}{l}{\textit{DeepLab}}\\[1pt]
P0 & 300 & 29,138 & 0.8298 & 0.8344 & 0.4446 & 0.4436\\
P1 & 200 & 18,599 & 0.8375 & 0.8420 & 0.4458 & 0.4548\\
P2 & 406 & 56,602 & 0.7472 & 0.7065 & 0.5519 & 0.6409\\
P3 & 400 & 49,934 & 0.7437 & 0.7072 & 0.5585 & 0.6333\\\addlinespace[3pt]
\multicolumn{7}{l}{\textit{OCRNet}}\\[1pt]
P0 & 300 & 32,610 & 0.8362 & 0.8345 & 0.4479 & 0.4454\\
P1 & 200 & 20,754 & 0.8389 & 0.8378 & 0.4528 & 0.4491\\
P2 & 406 & 42,317 & 0.7906 & 0.7558 & 0.5359 & 0.5932\\
P3 & 400 & 38,241 & 0.7879 & 0.7544 & 0.5335 & 0.5774\\\addlinespace[3pt]
\bottomrule\end{tabular*}\par\medskip
\textbf{(b) Matched selection at 90\% retention}\par\smallskip
\begin{tabular*}{\linewidth}{@{\extracolsep{\fill}}lrrrrrr@{}}\toprule
& \multicolumn{3}{c}{Retained failure (\%) $\downarrow$} & \multicolumn{3}{c}{Retained forecast LL $\downarrow$}\\
\cmidrule(lr){2-4}\cmidrule(lr){5-7}Population & Entropy & $p$ & $m$ & Entropy & $p$ & $m$\\\midrule
\multicolumn{7}{l}{\textit{SegFormer}}\\[1pt]
P0 & 36.846 & 33.814 & 34.961 & 0.4801 & 0.4986 & 0.4885\\
P1 & 38.948 & 36.034 & 37.052 & 0.4942 & 0.5139 & 0.5023\\
P2 & 55.712 & 53.687 & 53.941 & 0.5721 & 0.5969 & 0.5901\\
P3 & 56.099 & 54.375 & 54.481 & 0.5853 & 0.6062 & 0.6015\\\addlinespace[3pt]
\multicolumn{7}{l}{\textit{DeepLab}}\\[1pt]
P0 & 26.818 & 25.111 & 25.234 & 0.4188 & 0.4256 & 0.4246\\
P1 & 26.364 & 24.464 & 24.709 & 0.4206 & 0.4283 & 0.4265\\
P2 & 59.068 & 56.835 & 57.591 & 0.5376 & 0.5543 & 0.5531\\
P3 & 58.604 & 56.544 & 57.191 & 0.5450 & 0.5625 & 0.5611\\\addlinespace[3pt]
\multicolumn{7}{l}{\textit{OCRNet}}\\[1pt]
P0 & 29.647 & 27.457 & 27.803 & 0.4226 & 0.4336 & 0.4295\\
P1 & 28.846 & 26.749 & 27.173 & 0.4276 & 0.4365 & 0.4335\\
P2 & 51.357 & 48.132 & 49.143 & 0.5198 & 0.5400 & 0.5348\\
P3 & 51.457 & 48.176 & 49.162 & 0.5171 & 0.5387 & 0.5312\\\addlinespace[3pt]
\bottomrule\end{tabular*}
\endgroup
\caption{Absolute levels on all outcome-valid components. (a) AUROC pools components; CE is image-macro in nats; $N$ counts eligible components. (b) Retained failure (\%) and forecast LL (nats) at 90\%; LL evaluates fixed $p$. P0: source out-of-fold; P1/P2/P3: source-development/adverse-development/final populations.}\label{tab:primary-levels}
\end{table}

DeepLab/P1 makes the distinction between prediction and selection concrete. MetaSeg has higher pooled AUROC than $p$ (0.8420 versus 0.8375), with a favorable paired comparison, yet retains more failures at 90\% coverage (24.709\% versus 24.464\%). Its failure gain against $p$ is $-0.245$ percentage points, with interval $[-0.475,-0.004]$. Against entropy, the same ranking instead reduces failure by 1.655 points while increasing LL by 0.005819 nats. Pooled AUROC and image-macro retained failure differ in both functional and weighting; this example motivates direct evaluation of the intended selection rather than identifying a retention-boundary mechanism. Supplement S4 reports the complete predictive and selective comparisons.

The main trade-off persists when every comparator is restricted to nonempty interiors: all twelve failure intervals against entropy remain favorable and all twelve LL intervals remain adverse. Against direct $p$, no failure advantage is established under this restriction either (Supplement S4). Thus the entropy-referenced trade-off is not confined to empty-interior components. This sensitivity changes both the eligible population and its budget denominator.

\FloatBarrier
\subsection{Relearning existing fields reveals gains beyond the fitted score}
\begin{figure}[!htbp]\centering
\includegraphics[width=\linewidth]{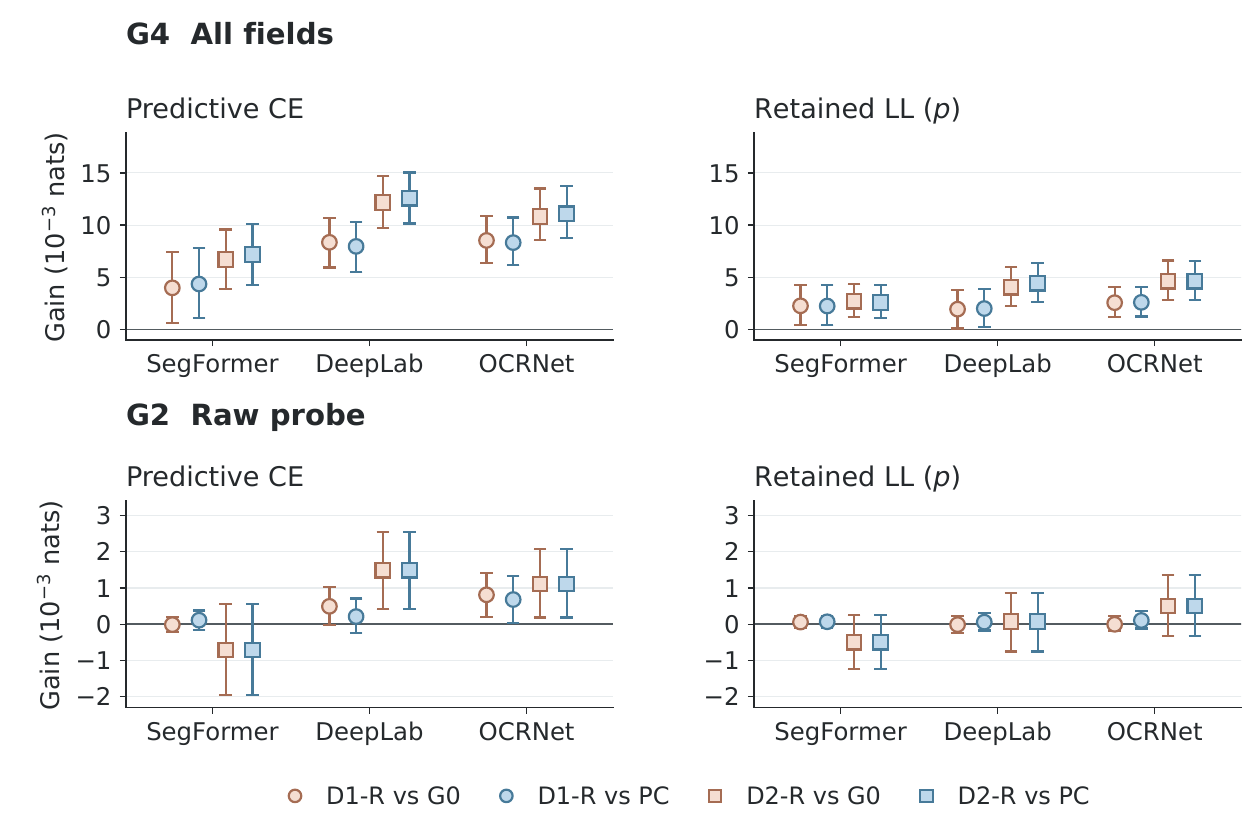}
\caption{Predictive and selective gains from G4 (all fields) and G2 (raw probe). Peach/blue denote learned $p$-only (G0) and dimension-matched $p$-derived controls (PC); circles/squares denote D1-R/D2-R. Positive favors added fields; bars show pointwise 95\% paired-image intervals. CE evaluates diagnostic forecasts; retained LL evaluates fixed $p$. Full values: Supplement S3, Tables S4--S5.}\label{fig:source}
\end{figure}

Limited selective gains from a fitted score do not determine what can be achieved with its existing inputs. We therefore relearn those fields under the matched source construction. Figure~\ref{fig:source} places predictive CE gains beside retained-LL gains: G4 uses all fields and G2 the raw probe, each compared with G0 and PC under both learners. Tables S4--S5 in Supplement S3 report the complete G1--G4 matrix.

Reusing all thirty-seven fields (G4) improves both prediction and selection in all three source systems. Under both learners, predictive CE and retained LL improve over G0 and PC (Fig.~\ref{fig:source}, top row). Retained LL also improves over entropy: the six gains range from 0.001967 to 0.003900 nats, all with positive intervals, although one lower bound is close to zero. These results identify a practical opportunity to improve reliability by relearning existing fields without adding visual inputs.

Predictive improvement need not establish a selection advantage. For the OCRNet probe (G2), both learners improve CE over G0 and PC, but every corresponding retained-LL interval crosses zero (Fig.~\ref{fig:source}, bottom row). Against PC, CE gains are 0.000682 and 0.001112 nats for D1-R and D2-R, whereas retained-LL gains are 0.000105 and 0.000501 nats. These small predictive gains leave the incremental selection advantage unresolved; the full intervals are in Supplement S3.

Gains vary with the field family, learner and architecture. For example, the probe's descriptors (G3) support predictive gains under both learners in DeepLab and OCRNet, with corresponding selective gains over G0 and entropy; SegFormer does not show the same pattern. The complete matrix in Supplement S3 includes favorable, adverse and unresolved results.

The source gains distinguish the value realized by a fitted score from the value attainable from its fields. In the original final-population evaluation, all nine frozen-score LL comparisons against forecast entropy were adverse (Supplement S6). The present controls show that useful predictive and selective gains can nevertheless be learned from existing fields on the source population. Because the population and nested forecast path differ, this establishes source opportunity, not recovery of the original target performance.

\FloatBarrier
\subsection{Benefits of loss-aligned remapping depend on eligibility and reference}
\begin{figure}[!t]\centering
\includegraphics[width=\linewidth]{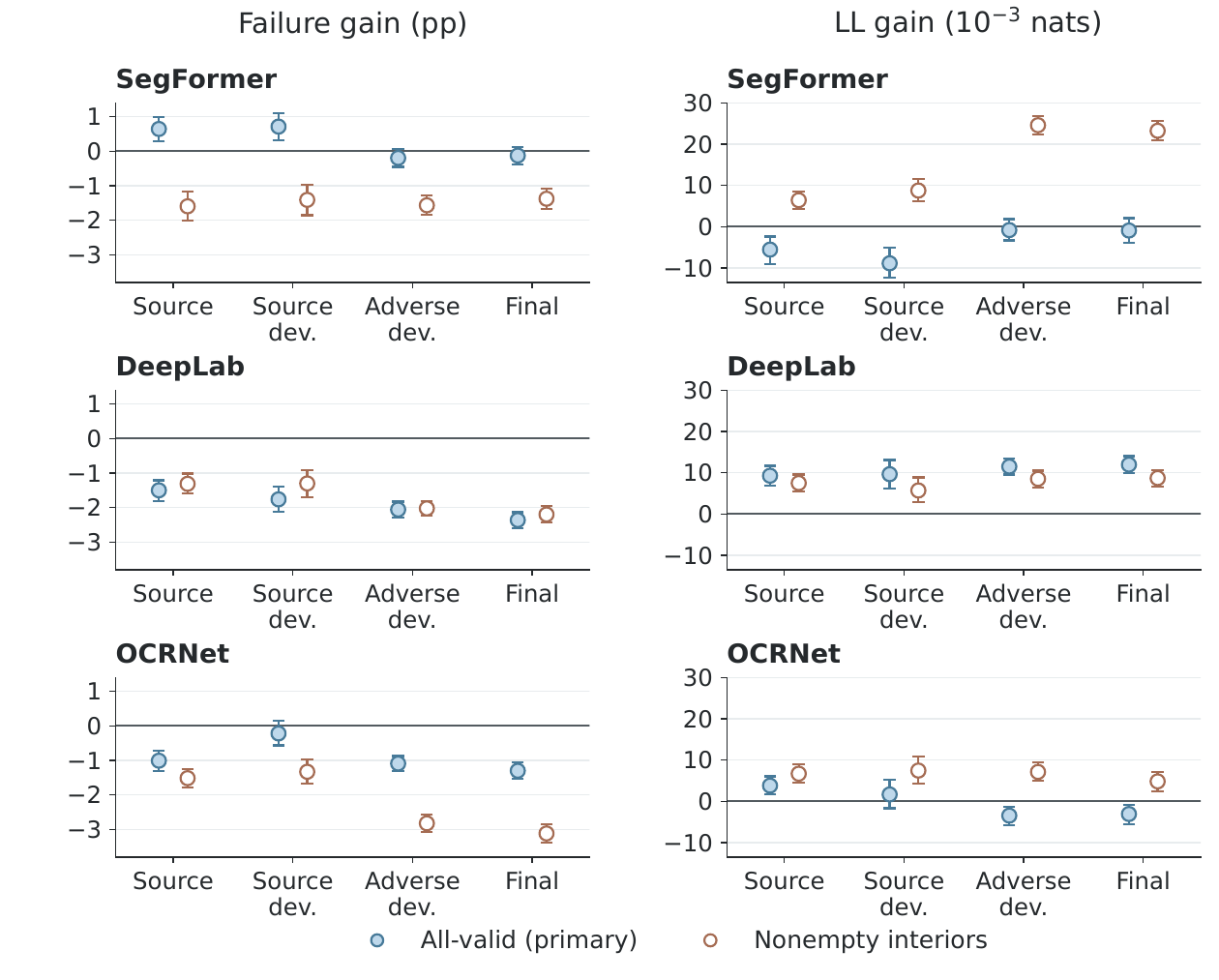}
\caption{Native-minus-mapped gains at 90\% retention; positive favors mapping. Filled blue circles denote all-valid components; open peach circles denote nonempty interiors, each with its own eligible set and matched budget. Bars show pointwise 95\% paired-image intervals. Forecast LL evaluates fixed $p$.}\label{fig:mapping}
\end{figure}

We next keep both forecasts fixed and change only the ranking rule, replacing native $m$ with $r_{\LL}(m,q)$. This tests whether aligning the score with forecast loss improves selection without refitting. On all-valid components, LL improves in DeepLab across all four populations and in OCRNet/P0, but worsens in SegFormer/P0--P1 and OCRNet/P2--P3; the remaining three intervals cross zero (Fig.~\ref{fig:mapping}, right column). Separately, comparison with entropy yields only two favorable LL intervals, DeepLab/P0--P1, and ten adverse intervals. Thus an appropriate risk target does not by itself establish a benefit when estimated probabilities are substituted. Exact comparisons are in Supplement S5, Tables S11--S12.

Lower LL can accompany more retained failures. In DeepLab/P1, mapping lowers LL from 0.426463 to 0.416817 nats while raising retained failure from 24.709\% to 26.470\%; the gains are $+0.009647$ nats and $-1.761$ percentage points. Across all-valid cells, failure intervals are two favorable, seven adverse and three crossing. Under nonempty-interior eligibility, all twelve LL gains are favorable (0.004834--0.024582 nats), yet failures increase in every cell. This is consistent with the different objectives: a correctly anticipated failure can incur low forecast loss while remaining a failed segmentation component. The two objectives require separate assessment.

\subsection{Exchanged components explain retained-loss changes}\label{sec:exchange-results}
Which components account for a ranking's gain or loss? At matched counts, shared components contribute equally to both selections. Let $A_j=S_j(s_0)\setminus S_j(s)$ and $B_j=S_j(s)\setminus S_j(s_0)$ be reference-only and candidate-only groups. Their losses give the exact difference
\begin{equation}\label{eq:exchange}
\Delta_{\ell,j}=\frac{\sum_{i\in A_j}\ell_{ji}-\sum_{i\in B_j}\ell_{ji}}{k_j}.
\end{equation}
The gain is therefore the removed loss minus the added loss, divided by the common budget. Supplement S5 further separates the exchanged fraction from the mean loss difference between the groups.

\FloatBarrier

\begin{figure}[!htbp]
\centering
\begin{minipage}[t]{.32\linewidth}\centering
\includegraphics[width=\linewidth]{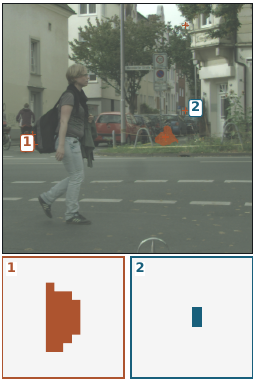}
\par\smallskip{\small (a) Source 1}
\end{minipage}\hfill
\begin{minipage}[t]{.32\linewidth}\centering
\includegraphics[width=\linewidth]{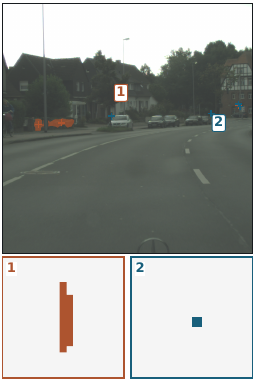}
\par\smallskip{\small (b) Source 2}
\end{minipage}\hfill
\begin{minipage}[t]{.32\linewidth}\centering
\includegraphics[width=\linewidth]{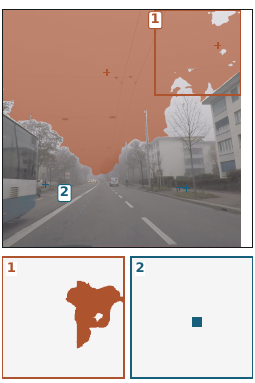}
\par\smallskip{\small (c) Fog}
\end{minipage}\par\medskip
\begin{minipage}[t]{.32\linewidth}\centering
\includegraphics[width=\linewidth]{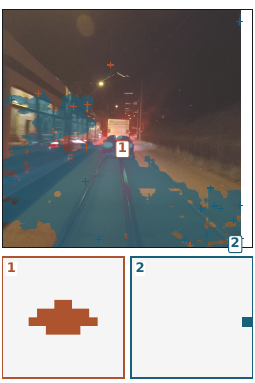}
\par\smallskip{\small (d) Night}
\end{minipage}\hfill
\begin{minipage}[t]{.32\linewidth}\centering
\includegraphics[width=\linewidth]{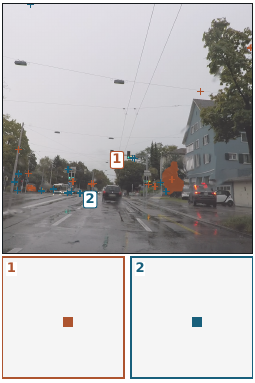}
\par\smallskip{\small (e) Rain}
\end{minipage}\hfill
\begin{minipage}[t]{.32\linewidth}\centering
\includegraphics[width=\linewidth]{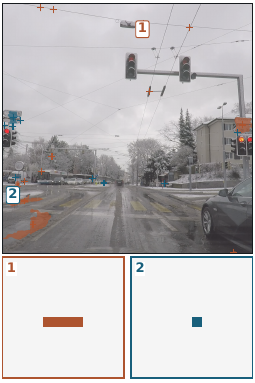}
\par\smallskip{\small (f) Snow}
\end{minipage}
\caption{Exchanged supports in six fixed DeepLab scenes at 90\% all-valid retention. Orange: native-only; blue: mapped-only. Crosses mark centroids. Paired labels locate mask insets, magnified independently. Selection uses the full image. Descriptive outcomes: Table~\ref{tab:case-levels}; scene and inset rules: Supplement S5.}\label{fig:case}
\end{figure}

Figure~\ref{fig:case} locates these exchanges in six DeepLab scenes already fixed for Figure~\ref{fig:components}: two source scenes and fog, night, rain and snow. Table~\ref{tab:case-levels} reports counts and outcomes. Source 2 improves LL by 0.020387 nats while increasing failure by 4.878 percentage points; Fog worsens LL by 0.031171 nats while reducing failure by 1.905 points. The scenes show how exchanged supports produce different endpoint directions; Figure~\ref{fig:mapping} supplies the population-level comparison.

\begin{table}[H]
\centering
{\fontsize{9.5}{11.5}\selectfont\setlength{\tabcolsep}{2pt}
\begin{tabular*}{\linewidth}{@{\extracolsep{\fill}}lrrrcccc@{}}
\toprule
& & & & \multicolumn{2}{c}{LL [nats]} & \multicolumn{2}{c}{Failures/$k$} \\
\cmidrule(lr){5-6}\cmidrule(lr){7-8}
Scene & $n$ & $k$ & $x$ & Native & Mapped & Native & Mapped \\
\midrule
Source 1 & 118 & 107 & 10 & 0.381016 & 0.397064 & 19/107 & 23/107 \\
Source 2 & 91 & 82 & 9 & 0.345995 & 0.325607 & 12/82 & 16/82 \\
Fog & 116 & 105 & 9 & 0.543883 & 0.575054 & 63/105 & 61/105 \\
Night & 241 & 217 & 22 & 0.541156 & 0.533147 & 164/217 & 169/217 \\
Rain & 178 & 161 & 17 & 0.535679 & 0.501192 & 83/161 & 89/161 \\
Snow & 212 & 191 & 21 & 0.629518 & 0.607752 & 95/191 & 96/191 \\
\bottomrule
\end{tabular*}}
\caption{Descriptive outcomes for the six scenes in Figure~\ref{fig:case} at 90\% all-valid retention. $n$: eligible count; $k$: retained count; $x$: removed and added count. LL is mean retained loss of fixed $p$ (nats); failures are counts over $k$.}\label{tab:case-levels}
\end{table}

\FloatBarrier

\begin{figure}[!htbp]
\centering
\includegraphics[width=\linewidth,trim=0 14.58bp 0 4.99bp,clip]{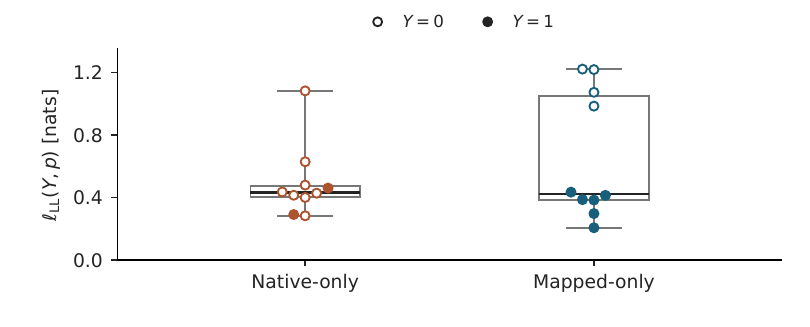}
\caption{Exclusive-component losses in Figure~\ref{fig:case}a: ten members per group, with failures filled. Boxes show medians and interquartile ranges; whiskers span minimum--maximum. Groups are unpaired and describe one image. Component identities: Supplement S5.}\label{fig:exchange-losses}
\end{figure}

Source 1, the original scene selected without viewing outcomes, illustrates the loss accounting in detail (Fig.~\ref{fig:exchange-losses}). Each ranking retains 107 components: 97 shared and ten exclusive to each group. The mapped-only supports span 1--46 pixels. Mapping replaces two failures with six, increasing retained failures by four; both endpoints favor native selection in this scene, unlike the population-level DeepLab/P1 LL result. The shared losses cancel. The difference depends on the sum of all exclusive losses, including high-loss members, rather than the group medians. The exclusive groups have no one-to-one pairing.

Changing many selections is not sufficient for a gain. Across the twelve primary cells, 98.5--100\% of images change retained identities, yet LL effects remain mixed. Equation~\eqref{eq:exchange} separates how much membership changes from the losses carried by those changes.

\FloatBarrier

\begin{figure}[!htbp]

\centering
\includegraphics[width=\linewidth,trim=0 14.83bp 0 13.02bp,clip]{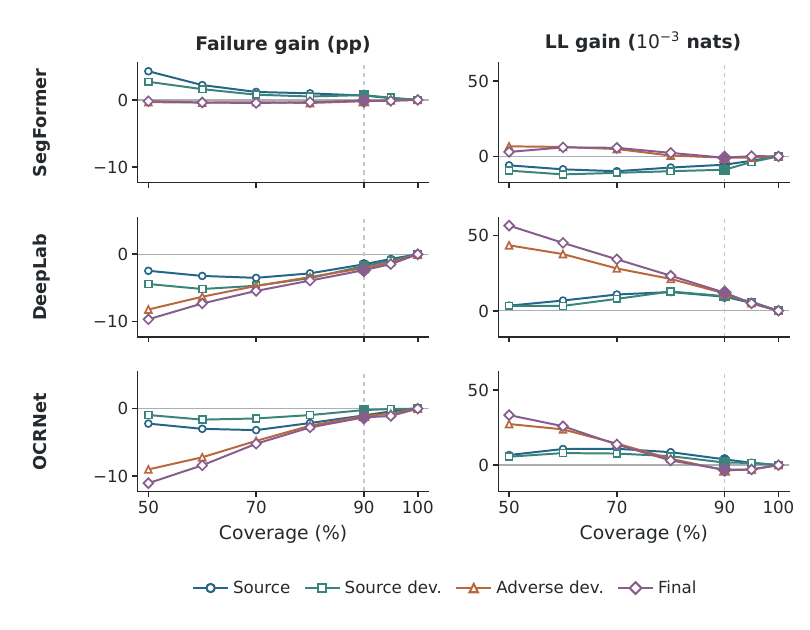}
\captionof{figure}{Native-minus-mapped gains across all-valid retention budgets; positive favors mapping. Columns show failure gain (percentage points) and LL gain ($10^{-3}$ nats). Filled markers denote 90\%; lines connect archived estimates. Colors denote P0--P3. Primary intervals: Figure~\ref{fig:mapping}.}\label{fig:coverage}

\end{figure}

\paragraph{Sensitivity to the retention budget}
The retention budget also changes the value of remapping (Fig.~\ref{fig:coverage}). All-valid LL point gains are positive in ten cells at each of 50--80\% retention, six at 90\% and seven at 95\%. At full retention, the sets coincide and all gains are zero. These archived curves describe budget sensitivity; 90\% remains the primary setting.

Whole-sequence resampling checks target-image dependence. For primary target LL, it leaves two favorable intervals, one adverse and three crossing; OCRNet/P2 changes from adverse under image resampling to crossing. Supplement S5 reports the sequence intervals and interior-only sensitivity.

\FloatBarrier

\FloatBarrier
\section{Discussion}
A component-reliability evaluation should inform a decision about the score, its inputs or its use. The observed reference dependence means that an additional score should be assessed against the ranking already available to the system. The source gains from G4 identify an opportunity to relearn existing fields, while the mixed mapping results call for checking the loss actually delivered by a new ranking rule. Matched comparisons make these decisions explicit.

\paragraph{From evaluation to diagnosis}
The combined evidence supports a more specific diagnosis than a predictive metric or a single ranking comparison. The available score can lack added screening value over $p$ while its existing fields still support source gains through G4 relearning. Separately, fixed-pair mapping locates the effect of changing selection with neither forecast refitted. These are distinct empirical comparisons, not successive stages of one learned selector.

Failure screening rewards fewer retained segmentation failures. Forecast auditing assesses the unchanged probability records on the selected components. Keeping the evaluated forecast fixed separates this selection effect from refitting or recalibration. When only screening is intended, failure is the relevant endpoint. When recorded-probability quality is also of interest, the audit reveals an additional, potentially competing consequence. The retained-set construction supports both uses while keeping their losses distinct.

The same-construction source comparison strengthens the evidence beyond an AUROC--selection contrast that also changes weighting. Its predictive and selective assessments share the source population, image nesting and control representations. G4 provides positive evidence on both endpoints; G2 establishes a predictive gain but leaves selection gains unresolved. The mechanism remains open: calibration error and ranking behavior near the retention boundary are possible explanations that these comparisons do not isolate.

\paragraph{Scope and further use}
The comparison design can guide other component-reliability studies by specifying the decision unit, budget and available reference, then separating the fitted score, reusable fields and ranking rule. The empirical findings here concern three systems on shared Cityscapes/ACDC resources, one MetaSeg classification adaptation and two source learner families. Base and MetaSeg field sets differ, and dimension-matched p-only controls do not equalize function classes. These gains establish a source-domain opportunity; transfer requires a separate evaluation. 

The outcome covers zero same-class overlap on predicted components; it excludes missed objects with no predicted component. Component-count budgets do not measure human review effort or downstream deployment benefit. Eligibility determines the selected population; sequence sensitivity covers mapping rather than every headline comparison. Historical outcome exposure and retrospective later P3 analyses limit confirmatory interpretation.

The supplement supplies the complete results, implementation specifications and analysis chronology. Portable synthetic tests check the matched evaluator; real-cache verification covers one comparison. Full upstream fitting and the complete research pipeline have not been independently regenerated. Supplement S8 separates document reproduction, evaluator checks and research-result reproduction.

\section{Conclusion}
We develop a decision-centered evaluation framework for component reliability around Predict, Realize and Select. It separately tests the value of an available score, the opportunity to relearn existing fields and the effect of changing the ranking rule. Across the evaluated systems, these comparisons reveal reference-dependent screening gains, constructive predictive and selective source gains from field reuse, and mixed effects of plug-in risk mapping. Matched budgets and exact membership accounting connect the findings to the components actually retained. Reporting these comparisons alongside predictive evidence provides a structured way to quantify and diagnose selective value.

\section*{CRediT authorship contribution statement}
\noindent\textbf{Minhui Kang:} Conceptualization, Methodology, Investigation, Formal analysis, Validation, Visualization, Writing -- original draft, Writing -- review \& editing, Supervision, Funding acquisition.

\noindent\textbf{Yiyao Zhang:} Conceptualization, Methodology, Software, Data curation, Investigation, Formal analysis, Validation, Visualization, Writing -- original draft, Writing -- review \& editing.

\noindent\textbf{Peizhe Long:} Software, Data curation, Validation.

\noindent\textbf{Jiafeng Qi:} Software, Visualization.

\noindent\textbf{Hongyu Jin:} Writing -- original draft, Writing -- review \& editing.

\noindent\textbf{Jiayun Zheng:} Writing -- original draft, Writing -- review \& editing.

\section*{Declaration of generative AI and AI-assisted technologies in the manuscript preparation process}
During the preparation of this work, the authors used \mbox{ChatGPT} and \mbox{OpenAI Codex} (OpenAI) for language editing and LaTeX typesetting. After using these tools, the authors reviewed and edited the content as needed and take full responsibility for the content of the published article.

\section*{Declaration of competing interests}
The authors declare no competing interests.

\section*{Funding}
This work was supported by the Science and Technology Research Project of the Jiangxi Provincial Department of Education, China (Grant No. GJJ2406408).

\section*{Data and code availability}
The evaluation code, experimental specifications, and numerical data supporting the reported results are available on Zenodo at \url{https://doi.org/10.5281/zenodo.23276040}. The archive does not redistribute the original datasets, model checkpoints, or population-wide component caches.
\setlength{\bibsep}{4pt plus 1pt minus 1pt}

\end{document}


\setlength{\LTcapwidth}{\textwidth}
\begin{frontmatter}
\title{Supplementary Material: From Failure Prediction to Component Selection in Semantic Segmentation}
\renewcommand{\thefnote}{$\dagger$}
\author[jx]{Minhui Kang\fnref{equal}\corref{cor1}}
\author[jx]{Yiyao Zhang\fnref{equal}}
\author[sd]{Peizhe Long}
\author[jx]{Jiafeng Qi}
\author[nchu]{Hongyu Jin}
\author[jx]{Jiayun Zheng}
\address[jx]{School of Power Engineering, Jiangxi Electric Vocational and Technical College, Nanchang, Jiangxi, China}
\address[sd]{College of Information Science and Engineering, Shandong Agricultural University, Tai'an, Shandong, China}
\address[nchu]{College of Aviation Manufacturing Industry, Nanchang Hangkong University, Jingdezhen, Jiangxi, China}
\fntext[equal]{These authors contributed equally to this work.}
\cortext[cor1]{Corresponding author: \href{mailto:zebukang@hotmail.com}{zebukang@hotmail.com}.}
\hypersetup{pdfauthor={Minhui Kang, Yiyao Zhang, Peizhe Long, Jiafeng Qi, Hongyu Jin, Jiayun Zheng}}

\begin{abstract}
This supplement provides implementation details and complete results for the comparisons in the main paper. Sections S1--S2 specify the evaluation and fitted models; S3--S5 give the field controls, MetaSeg comparisons and mapping sensitivities. Sections S6--S8 document the original frozen-score evaluation, additional analyses within this study and analysis chronology. S9 provides a literature map.
\end{abstract}
\end{frontmatter}

\section*{Guide to this supplement}
\noindent\begingroup\setlength{\tabcolsep}{4pt}\begin{tabular}{@{}p{.18\linewidth}p{.38\linewidth}p{.39\linewidth}@{}}
\textbf{Question} & \textbf{Main evidence} & \textbf{Supplemental detail}\\[3pt]
\textit{Predict} & Figure 4; Table 1 & S4: prediction and reference comparisons\\
\textit{Realize} & Figures 3, 5 & S2--S3: fitting and full field controls\\
\textit{Select} & Figures 6--9; Table 2 & S5: mapping and exchange sensitivities\\
\end{tabular}\par\endgroup\smallskip
For implementation and leakage checks, start with S2 and Algorithm S1. For complete numerical comparisons, use S3--S5. For the initial frozen-score evaluation, additional analyses and their timing, use S6--S8. Each section is accessible through PDF bookmarks.

\noindent\textbf{Table labels.} Architecture groups use SegFormer (A1), DeepLab (A2) and OCRNet (A3). Populations are Source (P0), Source dev.\ (P1), Adverse dev.\ (P2) and Final (P3). All and Interior denote all-valid and nonempty-interior eligibility. G0 is learned $p$-only; PC is the dimension-matched $p$-derived control. Labels retain the archived codes for lookup.

Table S1 identifies which checks accompany each comparison. Architecture--population cells share image resources. Sequence intervals are supplied for mapping on P2/P3, separately from the image intervals for other comparisons.

\begin{center}\small\setlength{\tabcolsep}{4pt}
\begin{tabular}{@{}p{105pt}p{78pt}p{86pt}p{65pt}p{93pt}@{}}\toprule
Analysis & Image intervals & Eligibility & Coverage & Sequence intervals\\\midrule
MetaSeg selection (S4) & Pointwise 95\% & All / interiors & 90\% & ---\\
Strict source controls (S3) & Pointwise 95\% & Source-valid & 90\% & ---\\
Fixed-pair mapping (S5) & Pointwise 95\% & All / interiors & 50--100\% grid & P2 / P3\\
Original frozen scores (S6) & Pointwise 95\% & Original valid & 90\% & ---\\\bottomrule
\end{tabular}
\captionof{table}{Scope of the available comparisons and sensitivities. A dash means that this check is not supplied for that analysis.}\label{tab:s-1}
\end{center}

\pdfbookmark[0]{S1. Evaluation definitions and selection}{supp-s1}
\section*{S1. Evaluation definitions and selection}\label{sec:s1}
The unit is an eight-connected predicted-class component $K_{ji}$ from a frozen segmentation map. Let $V_j$ exclude ignore-255 pixels and let $\mathcal G_{ji}$ contain the overlapping same-class ground-truth components. The binary outcome is
\[
Y_{ji}=\mathbf 1\!\left\{\max_{G\in\mathcal G_{ji}}
\frac{|(K_{ji}\cap V_j)\cap G|}{|(K_{ji}\cap V_j)\cup G|}=0\right\}.
\]
The maximum of an empty collection is zero. No outcome is assigned if $K_{ji}\cap V_j$ is empty. Feature generation is label-free; outcome-valid eligibility is an evaluation restriction. Identity is the tuple (architecture, population, image, component). Different architectures do not produce paired component identities.

For $n_j$ eligible components, each ranking retains $k_j=\lceil cn_j\rceil$. The implemented order is descending score and ascending integer component ID; the first $n_j-k_j$ items are rejected. This is not interchangeable at ties with ascending-score retention using the same ID direction. Empty eligible images are recorded and omitted from the image-macro loss; if all images are empty no estimate is returned. The registered coverage grid is 50, 60, 70, 80, 90, 95 and 100 percent.

The probabilities $p$ and $m$ forecast $Y=1$. The positive class of the fitted classifiers is class 1. R0, F1 and F3 are ranking scores, not probabilities. Retained LL always evaluates $p$, with only its logarithmic argument clipped to $q\in[10^{-6},1-10^{-6}]$. Scores are not clipped into probabilities. Within-image losses are averaged equally over evaluable images. A positive gain is reference loss minus candidate loss.

\subsection*{Risk identities}
For conditional failure probability $\eta$, the two risks are $r_{\rm fail}=\eta$ and
\[
r_{\rm LL}(\eta,q)=H(q)+(\eta-q)\log\frac{1-q}{q}.
\]
At fixed cardinality, exchanging a retained item with larger conditional risk for a rejected item with smaller risk reduces the expected sum. No outcome-independence assumption is needed. Substituting a fitted estimate for $\eta$ does not preserve this guarantee automatically. The original entropy reference uses the frozen convention for unmodified $p$; the clipped entropy in the identity is not silently substituted for it.

\subsection*{Synthetic retained-set exchange}
In a synthetic set retaining six of eight components, IDs 1--5 are shared. The reference retains ID 6 with $p=0.99$, $Y=1$ and LL rounded to 0.010; the candidate retains ID 7 with $p=0.50$, $Y=0$ and LL rounded to 0.693. Then $\Delta_{\rm fail}=(1-0)/6\approx+0.167$, whereas $\Delta_{\rm LL}\approx(0.010-0.693)/6\approx-0.114$ nats. The candidate is favored for failure and the reference for forecast loss. This synthetic exchange motivates the selection schematic in main Figure 1; it is not a study observation or the 90\% operating point.

\begin{figure}[H]\centering
\begin{minipage}[t]{.31\linewidth}\vspace{0pt}\includegraphics[width=\linewidth]{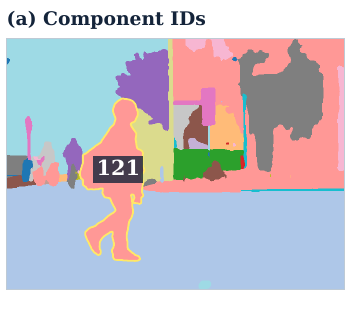}\end{minipage}\hspace{.05\linewidth}
\begin{minipage}[t]{.31\linewidth}\vspace{0pt}\includegraphics[width=\linewidth]{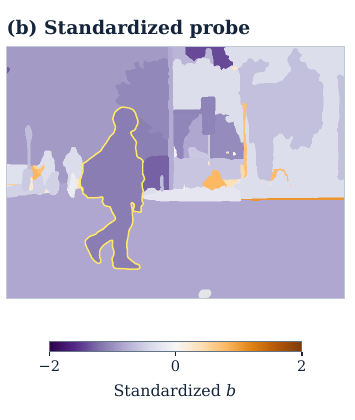}\end{minipage}
\caption{Component identities and probe output for main Figure 2a, with a wider crop and the same complete-image eligibility. (a) Repeating categorical colors encode identities, not classes or confidence; yellow marks component 121. (b) Source-standardized probe $b$ has a signed scale, not probability semantics. Gray marks excluded components.}\label{fig:s-component-probe}
\end{figure}

\pdfbookmark[0]{S2. Predictor and learner specifications}{supp-s2}
\section*{S2. Predictor and learner specifications}\label{sec:s2}
The specifications below describe the executed operators. P0, P1, P2 and P3 contain 300 source, 200 source-development, 406 adverse-development and 400 final images, respectively. A1--A3 denote SegFormer-B0, DeepLabV3 ResNetV1c-50-D8 and OCRNet HRNet-W48. The strict source comparison uses outer-clean base forecasts; the original P0 forecasts follow a separate fitting path.

\begin{table}[H]\centering
\textbf{Algorithm S1. Strict source construction.}\par\medskip
\small\begin{tabular}{@{}r p{.90\linewidth}@{}}\toprule
1 & Input the original five image folds and fixed upstream/downstream learner definitions.\\
2 & For each outer validation fold $V$, set $T$ to the other four folds.\\
3 & For each image fold $U\subset T$, fit upstream F2/F3 on $T\setminus U$, including preprocessing, and predict $U$.\\
4 & Join the inner-held-out rows to form downstream training features on $T$. All fits exclude labels from $V$.\\
5 & Fit upstream F2/F3 on all $T$ and generate outer-validation features on $V$.\\
6 & Fit each diagnostic's transformations and learner on the cross-fitted rows in $T$, using its specified image weights.\\
7 & Apply the frozen transformations and predictors to $V$.\\
8 & Use $Y_V$ to evaluate diagnostic CE. Map diagnostic probabilities to the LL selection score; evaluate retained LL on the outer-clean base forecast.\\
9 & Collect the five outer-validation outputs and evaluate G0, G1--G4, p-controls and R0 with the registered paired comparisons.\\\bottomrule
\end{tabular}
\end{table}

\pdfbookmark[1]{S2.1. Base forecast and fields}{supp-s21}
\subsection*{S2.1. Base failure forecast and the 37-field family}
\paragraph{Component geometry and pixel summaries}
Let $K$ be a predicted component, $B$ its base-predictor boundary and $I=K\setminus B$ its interior. A pixel belongs to $B$ if it lies at an image edge or has any four-neighbor outside $K$. For pixel-class probabilities $P_t(x)$, define
\[
h(x)=-\sum_t P_t(x)\log P_t(x),\qquad
u(x)=\max_t P_t(x),\qquad
g(x)=P_{(1)}(x)-P_{(2)}(x),
\]
where $P_{(1)}$ and $P_{(2)}$ are the largest and second-largest probabilities, and zero entropy terms are zero. The feature-mask size $N$ is $H_{\rm image}W_{\rm image}$ in the executed full-image calls (default all-true mask). The outcome mask excluding ignore-255 pixels is separate. An empty-interior mean is missing (\texttt{None}) in the raw table.

\needroom\begin{table}[H]\centering
\begingroup\fontsize{9}{11}\selectfont\setlength{\tabcolsep}{3pt}\renewcommand{\arraystretch}{1.02}
\begin{tabular}{p{111.0pt}p{111.0pt}p{111.0pt}p{111.0pt}}
\toprule
Index & Field & Formula & Unit \\
\midrule
1 & area & size(K) & pixels \\
2 & boundary\_\allowbreak{}size & size(B) & pixels \\
3 & interior\_\allowbreak{}size & size(I) & pixels \\
4 & mean\_\allowbreak{}max\_\allowbreak{}probability & mean\_\allowbreak{}K u & probability \\
5 & mean\_\allowbreak{}entropy & mean\_\allowbreak{}K h & nats \\
6 & mean\_\allowbreak{}probability\_\allowbreak{}margin & mean\_\allowbreak{}K g & probability difference \\
7 & mean\_\allowbreak{}predicted\_\allowbreak{}class\_\allowbreak{}probability & mean\_\allowbreak{}K P\_\allowbreak{}c & probability \\
8 & mean\_\allowbreak{}entropy\_\allowbreak{}boundary & mean\_\allowbreak{}B h & nats \\
9 & mean\_\allowbreak{}entropy\_\allowbreak{}interior & mean\_\allowbreak{}I h,\allowbreak{} missing if empty & nats \\
10 & mean\_\allowbreak{}max\_\allowbreak{}probability\_\allowbreak{}boundary & mean\_\allowbreak{}B u & probability \\
11 & mean\_\allowbreak{}max\_\allowbreak{}probability\_\allowbreak{}interior & mean\_\allowbreak{}I u,\allowbreak{} missing if empty & probability \\
12 & mean\_\allowbreak{}probability\_\allowbreak{}margin\_\allowbreak{}boundary & mean\_\allowbreak{}B g & probability difference \\
13 & mean\_\allowbreak{}probability\_\allowbreak{}margin\_\allowbreak{}interior & mean\_\allowbreak{}I g,\allowbreak{} missing if empty & probability difference \\
14 & relative\_\allowbreak{}component\_\allowbreak{}area & size(K)/\allowbreak{}max(N,\allowbreak{}1) & fraction \\
15 & relative\_\allowbreak{}boundary\_\allowbreak{}size & size(B)/\allowbreak{}max(size(K),\allowbreak{}1) & fraction \\
16 & relative\_\allowbreak{}interior\_\allowbreak{}size & size(I)/\allowbreak{}max(size(K),\allowbreak{}1) & fraction \\
17 & touches\_\allowbreak{}image\_\allowbreak{}boundary & indicator K intersects image edge & binary \\
18 & has\_\allowbreak{}interior & indicator size(I)\ensuremath{>}0 & binary \\
\bottomrule
\end{tabular}\endgroup
\caption{Ordered numerical fields of the base failure predictor. The nineteen class indicators follow these eighteen fields.}\label{tab:s-2}
\end{table}

Fields 19--37 are the predicted-class indicators $\mathbf1\{c=t\}$ for trainIds $t=0,\ldots,18$, in that order. They are concatenated unchanged after the eighteen numerical fields in Table S2. G4 reuses these raw inputs with a diagnostic learner. All inputs are label-free; outcome and supervision-validity fields are excluded.

\paragraph{Image-balanced preprocessing}
For $n_j$ valid supervised rows in image $j$, set $w_{ji}=1/n_j$. Fit each numerical mean to its finite training entries, using these weights; an all-missing column is an error. Fill missing or nonfinite entries with this mean. For the filled column $\widetilde x$, use
\[
\sigma^2=\frac{\sum_{ji}w_{ji}(\widetilde x_{ji}-\mu)^2}{\sum_{ji}w_{ji}},
\qquad x^*=(\widetilde x-\mu)/\sigma.
\]
An exactly constant column uses scale one. Apply the frozen means and scales at prediction; do not scale the class indicators.

\begin{specblock}{F2 base-forecast configuration.}
Classifier & \texttt{LogisticRegression}; L1 penalty; \texttt{liblinear} solver; $C=1$.\\
Intercept & Enabled; \texttt{intercept\_scaling}=1.\\
Optimization & Tolerance $10^{-4}$; maximum 5,000 iterations; random state 0.\\
Weights and checks & Image weights $1/n_j$; no class weights; both classes required; convergence warnings are errors.\\
Output & $p=\operatorname{sigmoid}(\theta_0+\boldsymbol\theta^\top x^*)$, probability of failure (class 1). No post-fit target calibration.\\
\end{specblock}

\pdfbookmark[1]{S2.2. Eleven descriptors}{supp-s22}
\subsection*{S2.2. Eleven component descriptors}
The ordered descriptor vector below is implemented by \texttt{matrix11}. Its exact field names are recorded in \path{polish/IMPLEMENTATION_CONTRACT.json}. Signs are learned by the downstream model.

\begin{center}\small\setlength{\tabcolsep}{4pt}
\begin{tabular}{@{}r p{.37\linewidth}p{.29\linewidth}p{.23\linewidth}@{}}\toprule
Index & Descriptor & Definition & Unit\\\midrule
1 & Log area & $\log |K|$ & Log pixel count\\
2 & Relative component area & $|K|/\max(N,1)$ & Fraction\\
3 & Relative boundary size & $|B|/\max(|K|,1)$ & Fraction\\
4 & Mean entropy & $\overline h_K$ & Nats\\
5 & Mean probability margin & $\overline g_K$ & Probability difference\\
6 & Boundary entropy & $\overline h_B$ & Nats\\
7 & Interior entropy & $\overline h_I$ & Nats\\
8 & Boundary margin & $\overline g_B$ & Probability difference\\
9 & Interior margin & $\overline g_I$ & Probability difference\\
10 & Geometry--uncertainty interaction & $\log|K|\,\overline h_K$ & Nats $\times$ log pixel count\\
11 & Boundary--uncertainty interaction & $\frac{|B|}{\max(|K|,1)}\overline h_B$ & Nats $\times$ fraction\\\bottomrule
\end{tabular}
\captionof*{table}{Descriptor order, definitions and units. Bars denote means over the indicated pixel set.}
\end{center}

\paragraph{Missing values and scaling}
Recompute log area from area before applying the specified finite fallback. A missing boundary or interior field uses its whole-component counterpart when present, otherwise zero; supplied nonfinite numbers become zero. For the F3 ridge, fit weighted medians and interquartile ranges to valid source-training rows with equal image mass and total weight one. Stable mergesort orders each column; the weighted quantile at $t$ is the first value whose cumulative mass is at least $t$ times total mass (left search). The scale is $\max(Q_{.75}-Q_{.25},10^{-6})$. Reuse these centers and scales at application. D1-R instead standardizes raw \texttt{matrix11} by weighted mean/SD; D2-R uses it raw.

\pdfbookmark[1]{S2.3. Calibration-opportunity probe}{supp-s23}
\subsection*{S2.3. Calibration-opportunity probe and frozen score}
\paragraph{Training-only calibration target}
Fit F2 within the authorized training partition and obtain its in-partition $p$. Set $x=\operatorname{logit}(\operatorname{clip}(p,10^{-6},1-10^{-6}))$. Fit a source-global affine probability $\sigma(a_gx+\beta_g)$ with normalized image-balanced logistic loss, and a separate $\sigma(a_jx+\beta_j)$ for each training image with uniform weights summing to one. Here $\sigma$ denotes sigmoid.

\begin{specblock}{Affine calibration optimizer, common to global and per-image fits.}
Method & L-BFGS-B with analytic gradient; initial $(a,\beta)=(1,0)$.\\
Bounds & $a\in[0.25,4]$; $\beta\in[-4,4]$; bound-hit tolerance $10^{-10}$.\\
Tolerances & \texttt{ftol}=$10^{-12}$; \texttt{gtol}=$10^{-8}$.\\
Limits & \texttt{maxiter}=5,000; \texttt{maxfun}=100,000; \texttt{maxls}=50.\\
Stable objective & $\operatorname{logaddexp}(0,ax+\beta)-Y(ax+\beta)$.\\
Fit records & Retain finite $(a,\beta)$ even at bounds or with nonconvergence flags; preserve those flags.\\
\end{specblock}

The component target is
\[
t_{ji}=\ell_{\rm LL}\!\left(Y_{ji},\sigma(a_gx_{ji}+\beta_g)\right)
       -\ell_{\rm LL}\!\left(Y_{ji},\sigma(a_jx_{ji}+\beta_j)\right),
\]
with target-loss probabilities clipped at $10^{-7}$. The image fit accesses training labels to construct this in-sample target. The target measures calibration opportunity relative to the global fit; it can be negative for individual components and differs from the observed LL of $p$ or an optimal rejection loss.

\paragraph{Ridge and score composition}
Fit normalized image-balanced squared error with slope penalty $\alpha=0.1$ and an unpenalized intercept. For the robustly scaled descriptors $\widetilde{\boldsymbol d}$ and design $X=[\boldsymbol1,\widetilde D]$,
\[
\boldsymbol\theta=(X^\top WX+\operatorname{diag}(0,\alpha,\ldots,\alpha))^{-1}X^\top Wt,
\qquad b=\theta_0+\widetilde{\boldsymbol d}^{\top}\boldsymbol\theta_{1:}.
\]
The raw probe $b$, affine offset $\beta$ and ridge coefficients $\boldsymbol\theta$ are distinct objects. Fit the weighted population mean/SD normalizers for $H(p)$, mean semantic entropy and raw $b$ on all label-free training rows, with equal image mass and equal component mass within each image. Replace zero SD by one. The frozen scores are
\[
R0=\operatorname{std}(H(p)),\quad
F1=\operatorname{std}(\overline h_K),\quad
F2=p,\quad
F3=R0+0.25\operatorname{std}(b).
\]
$H$ uses unmodified $p$ under the frozen entropy convention. Fold fits use their own normalizers; target deployment uses the full-source normalizers. F3 is a source-fitted ranking score with a calibration-opportunity target.

\pdfbookmark[1]{S2.4. Diagnostic learners}{supp-s24}
\subsection*{S2.4. Diagnostic learners}
G0 uses $p$ alone; G1 adds raw mean entropy, G2 raw probe $b$, G3 \texttt{matrix11} and G4 \texttt{matrix37}. Algorithm S1 defines the common source construction. The archived qualification covers 75 upstream fits and 240 downstream designs across five folds and three architectures.

\paragraph{D1-R design}
Let $u=\operatorname{logit}(q)$ for $q=\operatorname{clip}(p,10^{-6},1-10^{-6})$. Fit cubic quantile splines to raw $p$ with image weights $1/n_j$. Base columns are $u$ followed by spline outputs. Standardize additional $z$ with the training-only finite means, mean imputation and population SD of S2.1. Concatenate base columns, standardized $z$, then base-by-$z$ interactions in base-major order.

\begin{specblock}{D1-R learner configuration.}
Spline & \texttt{SplineTransformer}; degree 3; 8 knots; quantile knots; \texttt{include\_bias=False}. Other options retain the executed library defaults.\\
Classifier & \texttt{LogisticRegression}; L2 penalty; $C=1$; \texttt{lbfgs}; intercept enabled.\\
Optimization & Maximum 5,000 iterations; tolerance $10^{-8}$.\\
Weights & Image weights $1/n_j$; no class weights.\\
\end{specblock}

\paragraph{D2-R design}
Concatenate raw $p$ and raw $z$ (or $p$ alone), using native histogram missing-value handling and image weights $1/n_j$.

\begin{specblock}{D2-R learner configuration.}
Classifier & \texttt{HistGradientBoostingClassifier}; \texttt{loss=log\_loss}.\\
Tree and rate & Learning rate 0.05; 200 iterations; at most 15 leaves; unrestricted depth; minimum 50 samples per leaf.\\
Regularization & L2 regularization 1; early stopping disabled; random state 0.\\
Defaults & Remaining constructor arguments retain the executed environment's defaults; no added standardizer.\\
\end{specblock}

For both families, $\widehat\eta$ is class-1 probability clipped to $[10^{-6},1-10^{-6}]$. Parameters are fixed without target-data selection. Diagnostic CE evaluates $\widehat\eta$; retained LL ranks by $r_{\rm LL}(\widehat\eta,q)$ and evaluates the outer-clean $p$. CE references are learned G0 and the dimension-matched p-controls. LL comparisons additionally include R0, which rejects highest entropy. These diagnostics assess conditional source opportunity on P0.

\pdfbookmark[1]{S2.5. Dimension-matched p-derived controls}{supp-s25}
\subsection*{S2.5. Dimension-matched p-derived controls}
The dimension-matched $p$-derived controls (PC) use a fixed basis. For $q=\operatorname{clip}(p,10^{-6},1-10^{-6})$, its order is
\[
\boldsymbol\phi(p)=\bigl(p,p^2,p^3,H(q),\operatorname{logit}(q),
\sin(\pi p),\cos(\pi p),\sqrt p,\sqrt{1-p}\bigr).
\]
Zero-based column $j$ of the $d$-column control is $\phi_{j\bmod9}(p)$, $j=0,\ldots,d-1$. Thus $d=1$ uses $p$; $d=11$ adds $p,p^2$ after the nine terms; $d=37$ uses four cycles followed by $p$. G1/G2 pair with GCTRL1, G3 with GCTRL11 and G4 with GCTRL37. Add these pseudo-fields alongside $p$ under the corresponding diagnostic learner and folds. This matches column count under fixed p-derived transformations; it does not equalize function classes.

\pdfbookmark[1]{S2.6. MetaSeg adaptation}{supp-s26}
\subsection*{S2.6. MetaSeg classification adaptation}
\paragraph{Component boundary and features}
The adaptation follows MetaSeg's scalar/class-probability feature family and sparse logistic classification. Canonical component identities, areas and resolved maps are checked by hash. For MetaSeg, a pixel is boundary when any of its existing eight in-image neighbors belongs to another component; outside-image neighbors are omitted. This differs from the base predictor's edge-including, four-neighbor boundary. Let $S=|K|$, $S_{\rm in}=|I|$ and $S_{\rm bd}=|B|$; zero $S_{\rm bd}$ is an error.

Using float32 pixel probabilities, compute
\[
E(x)=\frac{\sum_t P_t(x)\log(P_t(x)+\epsilon_{\rm float32})}{\log(1/19)},
\qquad D(x)=1-P_{(1)}(x)+P_{(2)}(x).
\]
Heatmaps use float32 arithmetic; component sums use float64. For each stem $A\in\{E,D\}$, its five ordered statistics are
\[
\overline A_K,\quad\overline A_I,\quad\overline A_B,\quad
\overline A_K\frac{S}{S_{\rm bd}},\quad
\overline A_I\frac{S_{\rm in}}{S_{\rm bd}}.
\]
An empty-interior mean is zero here. The complete scalar order is the five $E$ terms, five $D$ terms, then $S,S_{\rm in},S_{\rm bd},S/S_{\rm bd},S_{\rm in}/S_{\rm bd}$. Append the nineteen class means $\overline{P_t}_K$, $t=0,\ldots,18$, for 34 fields. Sizes are pixel counts; the remaining quantities are dimensionless. Class means are probabilities, whereas G4 uses indicators. Exact field names and the upstream commit are in the implementation contract.

\begin{specblock}{MetaSeg adaptation: fitting and selection.}
Training rows & Valid P0 outcomes with $S_{\rm in}>0$.\\
Preprocessing & Unweighted component mean and population SD plus $10^{-10}$ for all 34 fields; re-estimate within each fit.\\
Classifier & \texttt{LogisticRegression}; L1; \texttt{saga}; maximum 1,000 iterations; tolerance $10^{-3}$; random state 0; no class weights; default intercept enabled; no image sample weights.\\
$C$ candidates & $10^e$ for $e=(-4.2,-2.95,-1.7,-0.45,0.8)$ in this order.\\
Selection & Maximize pooled component accuracy at probability threshold $\geq0.5$ in image-grouped inner folds. Exact ties select the first, smallest $C$.\\
P0 prediction & Original five outer image folds; select $C$ within the remaining four, fit their eligible rows, then score every valid outer component.\\
Deployment & Select $C$ by five-fold source validation; fit eligible full-P0 rows; freeze parameters for P1--P3.\\
Output & $m$ is class-1 failure probability, with higher scores rejected under the S1 tie rule.\\
\end{specblock}

Primary $E_{\rm ALL}$ contains every canonical valid-outcome component, including empty interiors. Sensitivity $E_{\rm IN}$ projects every comparator onto $S_{\rm in}>0$ and recomputes $k_j$. AUROC pools components; fixed-budget losses average images. The classifier adaptation uses source-only image nesting for this protocol; the original segmentation systems and adjusted-IoU regression are outside its scope.

\pdfbookmark[1]{S2.7. Specification summary}{supp-s27}
\subsection*{S2.7. Specification summary and code index}
Table S3 summarizes fitting and application. Exact field names, configurations and source-function hashes are indexed in \path{polish/IMPLEMENTATION_CONTRACT.json}; its paths refer to the research repository.

\needroom\begin{table}[H]\centering
\begingroup\small\setlength{\tabcolsep}{4pt}\renewcommand{\arraystretch}{1.15}
\textbf{(a) Inputs, targets and learners}\par\smallskip
\begin{tabular}{@{}p{.20\linewidth}p{.25\linewidth}p{.24\linewidth}p{.23\linewidth}@{}}\toprule
Object & Features & Target & Learner \\\midrule
Base forecast $p$ (F2) & 18 numeric + 19 indicators & Y & fixed L1 liblinear \\
Frozen score F3 & 11 descriptors & global-minus-image affine opportunity & robust-scaled ridge \\
Descriptors (G3) & p + raw 11 descriptors & Y & D1-R /\allowbreak{} D2-R \\
All fields (G4) & p + raw 37 fields & Y & D1-R /\allowbreak{} D2-R \\
MetaSeg & 15 E/\allowbreak{}D/\allowbreak{}S + 19 class means & Y & nested-C L1 saga \\
Fixed-pair mapping & available frozen (p,\allowbreak{}m) & no fit & deter\-ministic mapping \\
\bottomrule\end{tabular}
\medskip
\textbf{(b) Fitting, application and evaluation}\par\smallskip
\begin{tabular}{@{}p{.20\linewidth}p{.15\linewidth}p{.20\linewidth}p{.22\linewidth}p{.12\linewidth}@{}}\toprule
Object & Fit population & Validation / application & Ranking score & Evaluated forecast \\\midrule
Base forecast $p$ (F2) & P0 fold/\allowbreak{}full & original OOF /\allowbreak{} P1/\allowbreak{}P2/\allowbreak{}P3 & p & p \\
Frozen score F3 & P0 fold/\allowbreak{}full & original OOF /\allowbreak{} P1/\allowbreak{}P2/\allowbreak{}P3 & normalized H(p)+0.25 normalized b & p \\
Descriptors (G3) & fully nested P0 & outer P0 only & $r_{\mathrm{LL}}(\widehat{\eta},q)$ & outer-clean p \\
All fields (G4) & fully nested P0 & outer P0 only & $r_{\mathrm{LL}}(\widehat{\eta},q)$ & outer-clean p \\
MetaSeg & P0,\allowbreak{} S\_\allowbreak{}in\ensuremath{>}0 & P0 OOF; P1/\allowbreak{}P2/\allowbreak{}P3 full-source & m & base p (original fit) \\
Fixed-pair mapping & none & existing P0/\allowbreak{}P1/\allowbreak{}P2/\allowbreak{}P3 & $r_{\mathrm{LL}}(m,q)$ & same base p (original fit) \\
\bottomrule\end{tabular}
\endgroup
\caption{Model and ranking specifications. Each row in (b) corresponds to the same object in (a). Analysis timing and evidence status are given in S8.}\label{tab:s-3}
\end{table}

\pdfbookmark[0]{S3. Complete strict source controls}{supp-s3}
\section*{S3. Complete strict source controls}\label{sec:s3}
These controls implement Realize: they test gains attainable by relearning existing fields beyond a fitted use of the base probability. Diagnostic CE evaluates the learned forecast; retained LL evaluates selection of unchanged outer-clean base forecasts. Tables S4--S5 retain the complete G1--G4 comparisons; S2 specifies the common image-nested construction.

\noindent\textbf{Reading the two endpoints.} The same fitted diagnostic is assessed for prediction and for selection. Selection always evaluates the unchanged base forecast.
\begin{center}\small\setlength{\tabcolsep}{5pt}\renewcommand{\arraystretch}{1.15}
\begin{tabular}{@{}p{.19\linewidth}p{.37\linewidth}p{.37\linewidth}@{}}\toprule
 & \textbf{Predictive assessment} & \textbf{Selective assessment}\\\midrule
Components & All eligible validation components & Matched retained set of size $k_j$\\
Learned forecast & $\widehat\eta_z=h(p,z)$ & $r_{\rm LL}(\widehat\eta_z,q)$ supplies the ranking\\
Evaluated loss & CE of $\widehat\eta_z$ & LL of fixed $p$\\
References & G0 and PC forecasts & G0, PC and entropy rankings\\\bottomrule
\end{tabular}\end{center}
\noindent\textbf{Field and learner labels.} G1: mean entropy; G2: raw probe; G3: eleven descriptors; G4: all 37 fields. Spline and Boosting denote D1-R and D2-R. Table S4 covers prediction and selection against G0 and entropy; Table S5 adds the PC selection comparisons.

For D2-R, GCTRL1 repeats $p$ and the G1/G2 controls coincide with G0; their equal contrasts are not independent replications.

\needroom\begin{table}[H]\centering
\begingroup\fontsize{8.5}{10.5}\selectfont\setlength{\tabcolsep}{3pt}\renewcommand{\arraystretch}{1.02}
\begin{tabular}{p{49.29pt}p{40.37pt}p{85.58pt}p{85.58pt}p{85.58pt}p{85.58pt}}
\toprule
Fields & Learner & CE vs G0 & CE vs PC & LL vs G0 & LL vs Entropy \\
\midrule
\multicolumn{6}{l}{\textbf{SegFormer (A1)}}\\*
G1 & Spline & +0.004 \newline{}[-0.386,\allowbreak{} +0.371] & +0.126 \newline{}[-0.246,\allowbreak{} +0.487] & +0.086 \newline{}[-0.087,\allowbreak{} +0.272] & -0.028 \newline{}[-0.260,\allowbreak{} +0.202] \\
G1 & Boosting & -1.454 \newline{}[-2.627,\allowbreak{} -0.163] & -1.454 \newline{}[-2.627,\allowbreak{} -0.163] & -0.888 \newline{}[-1.718,\allowbreak{} -0.119] & -1.274 \newline{}[-2.191,\allowbreak{} -0.411] \\
G2 & Spline & -0.010 \newline{}[-0.222,\allowbreak{} +0.203] & +0.112 \newline{}[-0.150,\allowbreak{} +0.381] & +0.060 \newline{}[-0.099,\allowbreak{} +0.217] & -0.055 \newline{}[-0.267,\allowbreak{} +0.164] \\
G2 & Boosting & -0.708 \newline{}[-1.942,\allowbreak{} +0.558] & -0.708 \newline{}[-1.942,\allowbreak{} +0.558] & -0.493 \newline{}[-1.240,\allowbreak{} +0.251] & -0.878 \newline{}[-1.578,\allowbreak{} -0.168] \\
G3 & Spline & -0.043 \newline{}[-0.893,\allowbreak{} +0.838] & +0.174 \newline{}[-0.671,\allowbreak{} +1.105] & +0.013 \newline{}[-0.218,\allowbreak{} +0.262] & -0.101 \newline{}[-0.363,\allowbreak{} +0.171] \\
G3 & Boosting & +0.354 \newline{}[-1.442,\allowbreak{} +2.171] & +0.826 \newline{}[-1.065,\allowbreak{} +2.680] & -0.140 \newline{}[-1.406,\allowbreak{} +1.142] & -0.526 \newline{}[-1.735,\allowbreak{} +0.724] \\
G4 & Spline & +3.981 \newline{}[+0.634,\allowbreak{} +7.389] & +4.347 \newline{}[+1.098,\allowbreak{} +7.831] & +2.259 \newline{}[+0.438,\allowbreak{} +4.274] & +2.144 \newline{}[+0.338,\allowbreak{} +4.142] \\
G4 & Boosting & +6.709 \newline{}[+3.854,\allowbreak{} +9.567] & +7.181 \newline{}[+4.250,\allowbreak{} +10.092] & +2.698 \newline{}[+1.198,\allowbreak{} +4.320] & +2.313 \newline{}[+0.908,\allowbreak{} +3.818] \\
\multicolumn{6}{l}{\textbf{DeepLab (A2)}}\\*
G1 & Spline & +0.619 \newline{}[+0.149,\allowbreak{} +1.086] & +0.337 \newline{}[-0.090,\allowbreak{} +0.791] & +0.022 \newline{}[-0.108,\allowbreak{} +0.160] & +0.051 \newline{}[-0.089,\allowbreak{} +0.196] \\
G1 & Boosting & +0.964 \newline{}[-0.236,\allowbreak{} +2.086] & +0.964 \newline{}[-0.236,\allowbreak{} +2.086] & +0.683 \newline{}[-0.387,\allowbreak{} +1.780] & +0.344 \newline{}[-0.774,\allowbreak{} +1.432] \\
G2 & Spline & +0.494 \newline{}[-0.022,\allowbreak{} +1.031] & +0.212 \newline{}[-0.240,\allowbreak{} +0.712] & -0.010 \newline{}[-0.242,\allowbreak{} +0.237] & +0.018 \newline{}[-0.218,\allowbreak{} +0.271] \\
G2 & Boosting & +1.489 \newline{}[+0.413,\allowbreak{} +2.552] & +1.489 \newline{}[+0.413,\allowbreak{} +2.552] & +0.079 \newline{}[-0.751,\allowbreak{} +0.870] & -0.261 \newline{}[-1.017,\allowbreak{} +0.539] \\
G3 & Spline & +3.899 \newline{}[+2.547,\allowbreak{} +5.362] & +3.473 \newline{}[+2.062,\allowbreak{} +4.913] & +1.600 \newline{}[+0.637,\allowbreak{} +2.704] & +1.628 \newline{}[+0.681,\allowbreak{} +2.725] \\
G3 & Boosting & +7.353 \newline{}[+5.143,\allowbreak{} +9.699] & +7.771 \newline{}[+5.536,\allowbreak{} +10.131] & +2.802 \newline{}[+1.114,\allowbreak{} +4.587] & +2.462 \newline{}[+0.796,\allowbreak{} +4.306] \\
G4 & Spline & +8.345 \newline{}[+5.937,\allowbreak{} +10.660] & +7.954 \newline{}[+5.492,\allowbreak{} +10.296] & +1.938 \newline{}[+0.091,\allowbreak{} +3.788] & +1.967 \newline{}[+0.117,\allowbreak{} +3.835] \\
G4 & Boosting & +12.149 \newline{}[+9.712,\allowbreak{} +14.675] & +12.567 \newline{}[+10.152,\allowbreak{} +15.035] & +4.033 \newline{}[+2.223,\allowbreak{} +5.973] & +3.694 \newline{}[+1.827,\allowbreak{} +5.678] \\
\multicolumn{6}{l}{\textbf{OCRNet (A3)}}\\*
G1 & Spline & +0.609 \newline{}[+0.170,\allowbreak{} +1.020] & +0.478 \newline{}[+0.052,\allowbreak{} +0.902] & +0.012 \newline{}[-0.084,\allowbreak{} +0.109] & -0.142 \newline{}[-0.270,\allowbreak{} -0.024] \\
G1 & Boosting & -0.021 \newline{}[-1.030,\allowbreak{} +1.015] & -0.021 \newline{}[-1.030,\allowbreak{} +1.015] & -0.108 \newline{}[-0.835,\allowbreak{} +0.750] & -0.818 \newline{}[-1.584,\allowbreak{} -0.034] \\
G2 & Spline & +0.813 \newline{}[+0.203,\allowbreak{} +1.416] & +0.682 \newline{}[+0.045,\allowbreak{} +1.335] & -0.002 \newline{}[-0.197,\allowbreak{} +0.219] & -0.155 \newline{}[-0.374,\allowbreak{} +0.073] \\
G2 & Boosting & +1.112 \newline{}[+0.185,\allowbreak{} +2.070] & +1.112 \newline{}[+0.185,\allowbreak{} +2.070] & +0.501 \newline{}[-0.317,\allowbreak{} +1.367] & -0.209 \newline{}[-0.967,\allowbreak{} +0.554] \\
G3 & Spline & +3.898 \newline{}[+2.564,\allowbreak{} +5.327] & +3.631 \newline{}[+2.195,\allowbreak{} +5.199] & +1.227 \newline{}[+0.190,\allowbreak{} +2.450] & +1.073 \newline{}[+0.060,\allowbreak{} +2.274] \\
G3 & Boosting & +5.918 \newline{}[+3.812,\allowbreak{} +8.201] & +6.170 \newline{}[+4.016,\allowbreak{} +8.525] & +2.356 \newline{}[+0.763,\allowbreak{} +4.025] & +1.646 \newline{}[+0.021,\allowbreak{} +3.413] \\
G4 & Spline & +8.525 \newline{}[+6.399,\allowbreak{} +10.828] & +8.316 \newline{}[+6.162,\allowbreak{} +10.719] & +2.561 \newline{}[+1.188,\allowbreak{} +4.047] & +2.407 \newline{}[+1.046,\allowbreak{} +3.880] \\
G4 & Boosting & +10.824 \newline{}[+8.546,\allowbreak{} +13.494] & +11.076 \newline{}[+8.761,\allowbreak{} +13.757] & +4.610 \newline{}[+2.834,\allowbreak{} +6.609] & +3.900 \newline{}[+2.077,\allowbreak{} +5.978] \\
\bottomrule
\end{tabular}\endgroup
\caption{G1--G4 source gains with pointwise 95\% paired-image intervals, in $10^{-3}$ nats. CE compares against learned $p$-only (G0) and dimension-matched $p$-derived controls (PC); LL compares against G0 and entropy. Positive favors added fields. Table S5 completes the retained-LL controls.}\label{tab:s-4}
\end{table}

\needroom\begingroup\fontsize{8.5}{10.5}\selectfont\setlength{\tabcolsep}{3pt}\renewcommand{\arraystretch}{1.02}
\begin{longtable}{p{150.0pt}p{150.0pt}p{150.0pt}}
\toprule
Learner / fields & PC columns & control minus Gz LL [95\% CI] \\
\midrule
\endfirsthead
\multicolumn{3}{l}{\footnotesize Table \thetable\ (continued)}\\
\toprule
Learner / fields & PC columns & control minus Gz LL [95\% CI] \\
\midrule
\endhead

\multicolumn{3}{l}{\textbf{SegFormer (A1)}}\\*
Spline /\allowbreak{} G1 & 1 & +0.095 [-0.087,\allowbreak{} +0.289] \\*
Spline /\allowbreak{} G2 & 1 & +0.068 [-0.100,\allowbreak{} +0.232] \\*
Spline /\allowbreak{} G3 & 11 & +0.028 [-0.221,\allowbreak{} +0.278] \\*
Spline /\allowbreak{} G4 & 37 & +2.238 [+0.418,\allowbreak{} +4.260] \\*
Boosting /\allowbreak{} G1 & 1 & -0.888 [-1.718,\allowbreak{} -0.119] \\*
Boosting /\allowbreak{} G2 & 1 & -0.493 [-1.240,\allowbreak{} +0.251] \\*
Boosting /\allowbreak{} G3 & 11 & -0.263 [-1.581,\allowbreak{} +1.064] \\*
Boosting /\allowbreak{} G4 & 37 & +2.575 [+1.081,\allowbreak{} +4.225] \\
\multicolumn{3}{l}{\textbf{DeepLab (A2)}}\\*
Spline /\allowbreak{} G1 & 1 & +0.091 [-0.055,\allowbreak{} +0.238] \\*
Spline /\allowbreak{} G2 & 1 & +0.059 [-0.173,\allowbreak{} +0.311] \\*
Spline /\allowbreak{} G3 & 11 & +1.631 [+0.672,\allowbreak{} +2.733] \\*
Spline /\allowbreak{} G4 & 37 & +1.997 [+0.211,\allowbreak{} +3.841] \\*
Boosting /\allowbreak{} G1 & 1 & +0.683 [-0.387,\allowbreak{} +1.780] \\*
Boosting /\allowbreak{} G2 & 1 & +0.079 [-0.751,\allowbreak{} +0.870] \\*
Boosting /\allowbreak{} G3 & 11 & +3.188 [+1.500,\allowbreak{} +4.953] \\*
Boosting /\allowbreak{} G4 & 37 & +4.419 [+2.609,\allowbreak{} +6.359] \\
\multicolumn{3}{l}{\textbf{OCRNet (A3)}}\\*
Spline /\allowbreak{} G1 & 1 & +0.119 [-0.016,\allowbreak{} +0.256] \\*
Spline /\allowbreak{} G2 & 1 & +0.105 [-0.119,\allowbreak{} +0.359] \\*
Spline /\allowbreak{} G3 & 11 & +1.249 [+0.218,\allowbreak{} +2.471] \\*
Spline /\allowbreak{} G4 & 37 & +2.592 [+1.227,\allowbreak{} +4.074] \\*
Boosting /\allowbreak{} G1 & 1 & -0.108 [-0.835,\allowbreak{} +0.750] \\*
Boosting /\allowbreak{} G2 & 1 & +0.501 [-0.317,\allowbreak{} +1.367] \\*
Boosting /\allowbreak{} G3 & 11 & +2.356 [+0.769,\allowbreak{} +4.038] \\*
Boosting /\allowbreak{} G4 & 37 & +4.609 [+2.836,\allowbreak{} +6.575] \\
\bottomrule

\caption{Retained-LL gains against dimension-matched $p$-derived controls (PC), completing Table S4. All 24 field--learner--architecture comparisons are shown, in $10^{-3}$ nats with pointwise 95\% intervals.}\label{tab:s-5}\\
\end{longtable}\endgroup

\begin{figure}[H]\centering
\includegraphics[width=375pt]{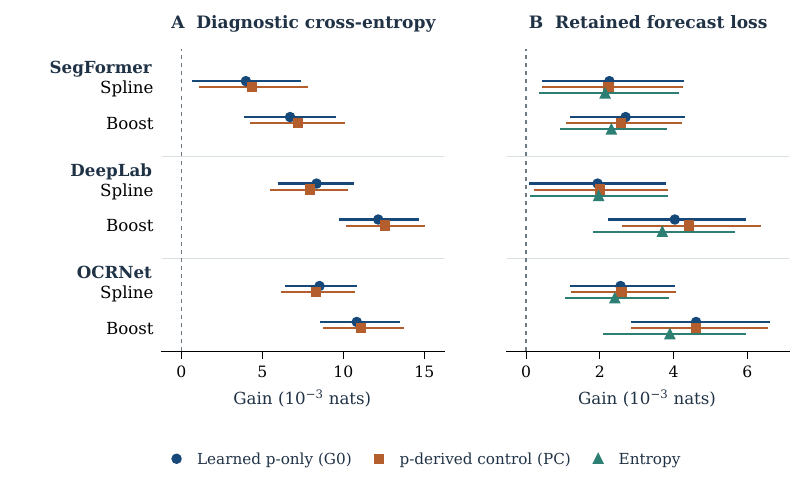}
\caption{Full-field (G4) gains against learned $p$-only, $p$-derived controls and entropy references, shown separately for each learner. Positive favors G4; bars show pointwise 95\% paired-image intervals.}\label{fig:s-source-controls}
\end{figure}
\pdfbookmark[0]{S4. MetaSeg prediction, selection and eligibility}{supp-s4}
\section*{S4. MetaSeg prediction, selection and eligibility}\label{sec:s4}
These comparisons connect Predict with Select by assessing the available MetaSeg score against both base-forecast references. S2.6 specifies the adaptation. The materialized evaluation contains 3,918 architecture--image cells from 1,306 unique images, yielding 459,415 label-free component instances, 381,175 valid outcomes and 311,297 valid nonempty interiors.

Native AUROC pools components; CE and retained losses are image-macro. The nonempty-interior analysis applies the same mask to every comparator and recomputes each within-image budget. This sensitivity changes the decision population and denominator.

\needroom\begingroup\fontsize{8.5}{10.5}\selectfont\setlength{\tabcolsep}{3pt}\renewcommand{\arraystretch}{1.02}
\begin{longtable}{p{90pt}p{42pt}p{75pt}p{75pt}p{75pt}p{75pt}}
\toprule
Population / eligibility & N components & MetaSeg AUROC & $p$ AUROC & MetaSeg CE & $p$ CE \\
\midrule
\endfirsthead
\multicolumn{6}{l}{\footnotesize Table \thetable\ (continued)}\\
\toprule
Population / eligibility & N components & MetaSeg AUROC & $p$ AUROC & MetaSeg CE & $p$ CE \\
\midrule
\endhead

\multicolumn{6}{l}{\textbf{SegFormer (A1)}}\\*
Source /\allowbreak{} All & 18728 & 0.715702 & 0.810513 & 0.607565 & 0.499287 \\*
Source /\allowbreak{} Interior & 15869 & 0.798600 & 0.803125 & 0.485110 & 0.483277 \\*
Src. dev. /\allowbreak{} All & 12209 & 0.736090 & 0.811533 & 0.592712 & 0.512086 \\*
Src. dev. /\allowbreak{} Interior & 10277 & 0.805113 & 0.803678 & 0.503272 & 0.501892 \\*
Adv. dev. /\allowbreak{} All & 32429 & 0.708956 & 0.756237 & 0.682546 & 0.583443 \\*
Adv. dev. /\allowbreak{} Interior & 25580 & 0.772327 & 0.766527 & 0.577341 & 0.588785 \\*
Final /\allowbreak{} All & 29614 & 0.697279 & 0.740307 & 0.688020 & 0.595325 \\*
Final /\allowbreak{} Interior & 23398 & 0.758212 & 0.752392 & 0.586377 & 0.600720 \\
\multicolumn{6}{l}{\textbf{DeepLab (A2)}}\\*
Source /\allowbreak{} All & 29138 & 0.834379 & 0.829771 & 0.443627 & 0.444571 \\*
Source /\allowbreak{} Interior & 25225 & 0.838801 & 0.829120 & 0.401928 & 0.415177 \\*
Src. dev. /\allowbreak{} All & 18599 & 0.842019 & 0.837550 & 0.454790 & 0.445788 \\*
Src. dev. /\allowbreak{} Interior & 16069 & 0.842332 & 0.833126 & 0.404630 & 0.414994 \\*
Adv. dev. /\allowbreak{} All & 56602 & 0.706512 & 0.747208 & 0.640949 & 0.551935 \\*
Adv. dev. /\allowbreak{} Interior & 44630 & 0.734393 & 0.769670 & 0.622537 & 0.544172 \\*
Final /\allowbreak{} All & 49934 & 0.707189 & 0.743655 & 0.633327 & 0.558502 \\*
Final /\allowbreak{} Interior & 39661 & 0.731530 & 0.764022 & 0.611800 & 0.551221 \\
\multicolumn{6}{l}{\textbf{OCRNet (A3)}}\\*
Source /\allowbreak{} All & 32610 & 0.834517 & 0.836193 & 0.445372 & 0.447894 \\*
Source /\allowbreak{} Interior & 27607 & 0.841877 & 0.835560 & 0.408237 & 0.418819 \\*
Src. dev. /\allowbreak{} All & 20754 & 0.837822 & 0.838872 & 0.449149 & 0.452838 \\*
Src. dev. /\allowbreak{} Interior & 17574 & 0.847783 & 0.841256 & 0.411086 & 0.420412 \\*
Adv. dev. /\allowbreak{} All & 42317 & 0.755820 & 0.790637 & 0.593182 & 0.535915 \\*
Adv. dev. /\allowbreak{} Interior & 34541 & 0.779622 & 0.803916 & 0.584595 & 0.525326 \\*
Final /\allowbreak{} All & 38241 & 0.754366 & 0.787907 & 0.577409 & 0.533462 \\*
Final /\allowbreak{} Interior & 30866 & 0.779789 & 0.802874 & 0.565758 & 0.521422 \\
\bottomrule

\caption{Native predictive performance under all-valid and nonempty-interior eligibility. CE is in nats.}\label{tab:s-6}\\
\end{longtable}\endgroup

\needroom\begingroup\fontsize{8.5}{10.5}\selectfont\setlength{\tabcolsep}{3pt}\renewcommand{\arraystretch}{1.02}
\begin{longtable}{p{228.0pt}p{228.0pt}}
\toprule
Population / eligibility & MetaSeg minus $p$: AUROC $\times1000$ [95\% CI] \\
\midrule
\endfirsthead
\multicolumn{2}{l}{\footnotesize Table \thetable\ (continued)}\\
\toprule
Population / eligibility & MetaSeg minus $p$: AUROC $\times1000$ [95\% CI] \\
\midrule
\endhead

\multicolumn{2}{l}{\textbf{SegFormer (A1)}}\\*
Source /\allowbreak{} All & -94.811 [-103.044,\allowbreak{} -86.755] \\*
Source /\allowbreak{} Interior & -4.525 [-7.242,\allowbreak{} -1.720] \\*
Src. dev. /\allowbreak{} All & -75.443 [-85.645,\allowbreak{} -65.133] \\*
Src. dev. /\allowbreak{} Interior & +1.435 [-2.501,\allowbreak{} +5.529] \\*
Adv. dev. /\allowbreak{} All & -47.281 [-54.037,\allowbreak{} -40.452] \\*
Adv. dev. /\allowbreak{} Interior & +5.800 [+2.576,\allowbreak{} +9.058] \\*
Final /\allowbreak{} All & -43.028 [-50.055,\allowbreak{} -35.275] \\*
Final /\allowbreak{} Interior & +5.820 [+2.621,\allowbreak{} +8.996] \\
\multicolumn{2}{l}{\textbf{DeepLab (A2)}}\\*
Source /\allowbreak{} All & +4.608 [+2.274,\allowbreak{} +7.068] \\*
Source /\allowbreak{} Interior & +9.680 [+7.156,\allowbreak{} +12.179] \\*
Src. dev. /\allowbreak{} All & +4.469 [+1.783,\allowbreak{} +7.280] \\*
Src. dev. /\allowbreak{} Interior & +9.207 [+6.017,\allowbreak{} +12.382] \\*
Adv. dev. /\allowbreak{} All & -40.696 [-45.088,\allowbreak{} -36.314] \\*
Adv. dev. /\allowbreak{} Interior & -35.278 [-39.898,\allowbreak{} -30.823] \\*
Final /\allowbreak{} All & -36.466 [-40.993,\allowbreak{} -31.901] \\*
Final /\allowbreak{} Interior & -32.492 [-37.002,\allowbreak{} -27.847] \\
\multicolumn{2}{l}{\textbf{OCRNet (A3)}}\\*
Source /\allowbreak{} All & -1.676 [-4.015,\allowbreak{} +0.858] \\*
Source /\allowbreak{} Interior & +6.317 [+4.548,\allowbreak{} +8.175] \\*
Src. dev. /\allowbreak{} All & -1.050 [-3.333,\allowbreak{} +1.342] \\*
Src. dev. /\allowbreak{} Interior & +6.527 [+4.613,\allowbreak{} +8.578] \\*
Adv. dev. /\allowbreak{} All & -34.817 [-39.083,\allowbreak{} -30.338] \\*
Adv. dev. /\allowbreak{} Interior & -24.294 [-28.079,\allowbreak{} -20.599] \\*
Final /\allowbreak{} All & -33.541 [-37.964,\allowbreak{} -29.105] \\*
Final /\allowbreak{} Interior & -23.085 [-27.076,\allowbreak{} -19.071] \\
\bottomrule

\caption{Paired MetaSeg-minus-$p$ AUROC differences, with 95\% image-cluster bootstrap intervals.}\label{tab:s-7}\\
\end{longtable}\endgroup

Table S6 gives absolute predictive performance; Table S7 gives paired AUROC differences using 2,000 image-cluster bootstrap draws with registered seed 2026092502. Table S8 gives retained-loss comparisons against entropy and direct probability, using the original matched-comparison stream shown in main Figure 4. In Table S8, LL versus entropy and failure versus direct probability were the original task-matched comparisons; the crossed references were completed retrospectively. Tables S9--S10 retain the separate eligibility-audit streams.

\needroom\begin{table}[H]\centering
\begingroup\fontsize{8.5}{10.5}\selectfont\setlength{\tabcolsep}{3pt}\renewcommand{\arraystretch}{1.02}
\begin{tabular}{p{47.57pt}p{97.61pt}p{97.61pt}p{97.61pt}p{97.61pt}}
\toprule
Population & LL vs entropy & LL vs $p$ & Failure vs entropy & Failure vs $p$ \\
\midrule
\multicolumn{5}{l}{\textbf{SegFormer (A1)}}\\*
Source & -8.404 \newline{}[-10.368,\allowbreak{} -6.455] & +10.100 \newline{}[+7.773,\allowbreak{} +12.596] & +1.885 \newline{}[+1.511,\allowbreak{} +2.252] & -1.147 \newline{}[-1.414,\allowbreak{} -0.866] \\
Src. dev. & -8.010 \newline{}[-10.215,\allowbreak{} -5.774] & +11.683 \newline{}[+8.312,\allowbreak{} +15.035] & +1.896 \newline{}[+1.426,\allowbreak{} +2.340] & -1.018 \newline{}[-1.353,\allowbreak{} -0.679] \\
Adv. dev. & -18.003 \newline{}[-19.594,\allowbreak{} -16.344] & +6.758 \newline{}[+4.350,\allowbreak{} +9.058] & +1.771 \newline{}[+1.501,\allowbreak{} +2.032] & -0.254 \newline{}[-0.478,\allowbreak{} -0.031] \\
Final & -16.282 \newline{}[-17.984,\allowbreak{} -14.515] & +4.607 \newline{}[+1.878,\allowbreak{} +7.553] & +1.617 \newline{}[+1.360,\allowbreak{} +1.888] & -0.106 \newline{}[-0.345,\allowbreak{} +0.137] \\
\multicolumn{5}{l}{\textbf{DeepLab (A2)}}\\*
Source & -5.869 \newline{}[-7.614,\allowbreak{} -4.119] & +0.941 \newline{}[-0.292,\allowbreak{} +2.132] & +1.583 \newline{}[+1.294,\allowbreak{} +1.879] & -0.124 \newline{}[-0.320,\allowbreak{} +0.059] \\
Src. dev. & -5.819 \newline{}[-7.888,\allowbreak{} -3.658] & +1.884 \newline{}[+0.322,\allowbreak{} +3.573] & +1.655 \newline{}[+1.282,\allowbreak{} +2.010] & -0.245 \newline{}[-0.475,\allowbreak{} -0.004] \\
Adv. dev. & -15.439 \newline{}[-17.046,\allowbreak{} -13.835] & +1.272 \newline{}[-0.805,\allowbreak{} +3.392] & +1.478 \newline{}[+1.245,\allowbreak{} +1.699] & -0.756 \newline{}[-0.937,\allowbreak{} -0.584] \\
Final & -16.065 \newline{}[-17.892,\allowbreak{} -14.411] & +1.413 \newline{}[-0.564,\allowbreak{} +3.552] & +1.412 \newline{}[+1.193,\allowbreak{} +1.637] & -0.647 \newline{}[-0.821,\allowbreak{} -0.460] \\
\multicolumn{5}{l}{\textbf{OCRNet (A3)}}\\*
Source & -6.951 \newline{}[-8.929,\allowbreak{} -4.964] & +4.088 \newline{}[+2.283,\allowbreak{} +5.941] & +1.843 \newline{}[+1.549,\allowbreak{} +2.145] & -0.347 \newline{}[-0.531,\allowbreak{} -0.171] \\
Src. dev. & -5.913 \newline{}[-8.280,\allowbreak{} -3.546] & +3.026 \newline{}[+1.103,\allowbreak{} +4.997] & +1.673 \newline{}[+1.292,\allowbreak{} +2.039] & -0.424 \newline{}[-0.666,\allowbreak{} -0.184] \\
Adv. dev. & -14.970 \newline{}[-16.662,\allowbreak{} -13.145] & +5.159 \newline{}[+2.524,\allowbreak{} +7.691] & +2.214 \newline{}[+1.971,\allowbreak{} +2.473] & -1.011 \newline{}[-1.214,\allowbreak{} -0.804] \\
Final & -14.149 \newline{}[-16.079,\allowbreak{} -12.085] & +7.405 \newline{}[+4.863,\allowbreak{} +9.799] & +2.295 \newline{}[+2.043,\allowbreak{} +2.538] & -0.986 \newline{}[-1.191,\allowbreak{} -0.785] \\
\bottomrule
\end{tabular}\endgroup
\caption{MetaSeg gains from the original matched-analysis stream (main Figure 4). LL: $10^{-3}$ nats; failure: percentage points. Positive favors MetaSeg; brackets show pointwise 95\% intervals.}\label{tab:s-8}
\end{table}

\subsection*{Complete eligibility sensitivities}
Tables S9--S10 report retrospective, score-preserving sensitivities for both references and both losses under each eligibility set, with the associated absolute retained losses. Their registered eligibility-audit stream uses 2,000 draws and seed 2026092501. All-valid points agree with Table S8, but bootstrap quantiles can differ across the two registered streams. Each interval remains attached to its own source comparison. The complete nonempty-interior analysis retains twelve favorable failure and twelve adverse LL intervals against entropy. Against direct probability, failure has five adverse and seven crossing intervals, and LL has six favorable and six crossing intervals.
\begin{table}[H]\centering\renewcommand{\arraystretch}{1.0}
\subsection*{Complete eligibility sensitivity: All-valid components}
\begin{center}\fontsize{9}{10}\selectfont\setlength{\tabcolsep}{5pt}
\begin{tabular}{@{}llrr@{}}\toprule
Population & Reference & Failure gain [95\% CI] & LL gain [95\% CI]\\\midrule
\multicolumn{4}{l}{\textbf{SegFormer (A1)}}\\*
Source & Entropy & 1.885 [1.530, 2.251] & -8.404 [-10.356, -6.360]\\
 & Probability & -1.147 [-1.428, -0.873] & 10.100 [7.635, 12.470]\\
Src. dev. & Entropy & 1.896 [1.436, 2.331] & -8.010 [-10.287, -5.819]\\
 & Probability & -1.018 [-1.342, -0.657] & 11.683 [8.309, 14.798]\\
Adv. dev. & Entropy & 1.771 [1.504, 2.044] & -18.003 [-19.515, -16.347]\\
 & Probability & -0.254 [-0.469, -0.025] & 6.758 [4.203, 9.022]\\
Final & Entropy & 1.617 [1.362, 1.885] & -16.282 [-17.905, -14.522]\\
 & Probability & -0.106 [-0.352, 0.123] & 4.607 [1.723, 7.503]\\
\addlinespace
\multicolumn{4}{l}{\textbf{DeepLab (A2)}}\\*
Source & Entropy & 1.583 [1.309, 1.865] & -5.869 [-7.588, -4.130]\\
 & Probability & -0.124 [-0.315, 0.060] & 0.941 [-0.214, 2.106]\\
Src. dev. & Entropy & 1.655 [1.275, 2.037] & -5.819 [-8.014, -3.796]\\
 & Probability & -0.245 [-0.482, -0.005] & 1.884 [0.275, 3.534]\\
Adv. dev. & Entropy & 1.478 [1.237, 1.718] & -15.439 [-17.092, -13.668]\\
 & Probability & -0.756 [-0.940, -0.581] & 1.272 [-0.748, 3.540]\\
Final & Entropy & 1.412 [1.194, 1.630] & -16.065 [-17.737, -14.300]\\
 & Probability & -0.647 [-0.827, -0.480] & 1.413 [-0.477, 3.493]\\
\addlinespace
\multicolumn{4}{l}{\textbf{OCRNet (A3)}}\\*
Source & Entropy & 1.843 [1.543, 2.143] & -6.951 [-8.915, -4.862]\\
 & Probability & -0.347 [-0.516, -0.169] & 4.088 [2.233, 5.925]\\
Src. dev. & Entropy & 1.673 [1.292, 2.067] & -5.913 [-8.207, -3.505]\\
 & Probability & -0.424 [-0.670, -0.177] & 3.026 [0.919, 5.038]\\
Adv. dev. & Entropy & 2.214 [1.966, 2.452] & -14.970 [-16.831, -13.065]\\
 & Probability & -1.011 [-1.211, -0.802] & 5.159 [2.631, 7.684]\\
Final & Entropy & 2.295 [2.043, 2.530] & -14.149 [-16.100, -12.263]\\
 & Probability & -0.986 [-1.188, -0.783] & 7.405 [4.919, 9.727]\\
\bottomrule\end{tabular}\end{center}
\noindent Absolute retained levels for the same rows (failure in percent; LL in nats). These are image-macro levels, not pooled-component prevalence.
\begin{center}\fontsize{9}{10}\selectfont\setlength{\tabcolsep}{5pt}
\begin{tabular}{@{}lrrrrrr@{}}\toprule
& \multicolumn{2}{c}{Entropy} & \multicolumn{2}{c}{Probability} & \multicolumn{2}{c}{MetaSeg}\\
Population & Failure & LL & Failure & LL & Failure & LL\\\midrule
\multicolumn{7}{l}{\textbf{SegFormer (A1)}}\\*
Source & 36.846 & 0.480 & 33.814 & 0.499 & 34.961 & 0.488\\
Src. dev. & 38.948 & 0.494 & 36.034 & 0.514 & 37.052 & 0.502\\
Adv. dev. & 55.712 & 0.572 & 53.687 & 0.597 & 53.941 & 0.590\\
Final & 56.099 & 0.585 & 54.375 & 0.606 & 54.481 & 0.602\\
\addlinespace
\multicolumn{7}{l}{\textbf{DeepLab (A2)}}\\*
Source & 26.818 & 0.419 & 25.111 & 0.426 & 25.234 & 0.425\\
Src. dev. & 26.364 & 0.421 & 24.464 & 0.428 & 24.709 & 0.426\\
Adv. dev. & 59.068 & 0.538 & 56.835 & 0.554 & 57.591 & 0.553\\
Final & 58.604 & 0.545 & 56.544 & 0.562 & 57.191 & 0.561\\
\addlinespace
\multicolumn{7}{l}{\textbf{OCRNet (A3)}}\\*
Source & 29.647 & 0.423 & 27.457 & 0.434 & 27.803 & 0.430\\
Src. dev. & 28.846 & 0.428 & 26.749 & 0.436 & 27.173 & 0.433\\
Adv. dev. & 51.357 & 0.520 & 48.132 & 0.540 & 49.143 & 0.535\\
Final & 51.457 & 0.517 & 48.176 & 0.539 & 49.162 & 0.531\\
\bottomrule\end{tabular}\end{center}
\caption{All-valid components at 90\% retention. Gains are reference minus MetaSeg: failure in percentage points; LL in $10^{-3}$ nats. Brackets show pointwise 95\% paired-image intervals. Comparators share eligible IDs and the recomputed budget within each row.}\label{tab:s-9}
\noindent\begin{minipage}{\linewidth}\footnotesize\raggedright
\noindent Source: \texttt{ELIGIBILITY\_MATCHED\_RESULTS.json}; 2,000 draws, seed 2026092501. Intervals use the separate eligibility-audit stream described in S4.
\end{minipage}
\end{table}

\begin{table}[H]\centering\renewcommand{\arraystretch}{1.0}
\subsection*{Complete eligibility sensitivity: Nonempty-interior components}
\begin{center}\fontsize{9}{10}\selectfont\setlength{\tabcolsep}{5pt}
\begin{tabular}{@{}llrr@{}}\toprule
Population & Reference & Failure gain [95\% CI] & LL gain [95\% CI]\\\midrule
\multicolumn{4}{l}{\textbf{SegFormer (A1)}}\\*
Source & Entropy & 2.075 [1.675, 2.473] & -9.217 [-11.313, -7.129]\\
 & Probability & -0.289 [-0.506, -0.081] & 1.776 [0.667, 2.988]\\
Src. dev. & Entropy & 1.721 [1.256, 2.162] & -8.177 [-10.676, -5.758]\\
 & Probability & -0.141 [-0.400, 0.122] & 1.336 [-0.306, 2.989]\\
Adv. dev. & Entropy & 1.850 [1.557, 2.141] & -19.219 [-20.954, -17.398]\\
 & Probability & -0.064 [-0.219, 0.100] & 2.337 [1.056, 3.577]\\
Final & Entropy & 1.685 [1.416, 1.960] & -17.948 [-19.748, -16.023]\\
 & Probability & -0.002 [-0.187, 0.173] & 1.734 [0.374, 3.160]\\
\addlinespace
\multicolumn{4}{l}{\textbf{DeepLab (A2)}}\\*
Source & Entropy & 1.416 [1.127, 1.694] & -3.328 [-4.790, -1.867]\\
 & Probability & 0.125 [-0.058, 0.318] & 0.343 [-0.714, 1.431]\\
Src. dev. & Entropy & 1.330 [0.990, 1.706] & -2.199 [-4.187, -0.306]\\
 & Probability & 0.175 [-0.051, 0.427] & 0.674 [-0.801, 2.378]\\
Adv. dev. & Entropy & 1.187 [0.943, 1.431] & -13.667 [-15.226, -12.094]\\
 & Probability & -1.006 [-1.226, -0.792] & 2.874 [0.810, 4.936]\\
Final & Entropy & 0.855 [0.610, 1.096] & -14.290 [-16.005, -12.611]\\
 & Probability & -1.085 [-1.307, -0.866] & 3.627 [1.526, 5.692]\\
\addlinespace
\multicolumn{4}{l}{\textbf{OCRNet (A3)}}\\*
Source & Entropy & 1.643 [1.364, 1.920] & -4.413 [-6.266, -2.482]\\
 & Probability & 0.079 [-0.061, 0.223] & 0.754 [-0.499, 2.164]\\
Src. dev. & Entropy & 1.548 [1.170, 1.967] & -4.739 [-7.180, -2.340]\\
 & Probability & -0.152 [-0.337, 0.028] & 0.858 [-0.351, 2.073]\\
Adv. dev. & Entropy & 2.004 [1.739, 2.263] & -14.938 [-16.782, -12.896]\\
 & Probability & -0.847 [-1.057, -0.653] & 1.175 [-1.273, 3.702]\\
Final & Entropy & 2.214 [1.951, 2.472] & -14.359 [-16.459, -12.309]\\
 & Probability & -0.701 [-0.916, -0.467] & 3.579 [0.774, 6.106]\\
\bottomrule\end{tabular}\end{center}
\noindent Absolute retained levels for the same rows (failure in percent; LL in nats). These are image-macro levels, not pooled-component prevalence.
\begin{center}\fontsize{9}{10}\selectfont\setlength{\tabcolsep}{5pt}
\begin{tabular}{@{}lrrrrrr@{}}\toprule
& \multicolumn{2}{c}{Entropy} & \multicolumn{2}{c}{Probability} & \multicolumn{2}{c}{MetaSeg}\\
Population & Failure & LL & Failure & LL & Failure & LL\\\midrule
\multicolumn{7}{l}{\textbf{SegFormer (A1)}}\\*
Source & 30.854 & 0.463 & 28.490 & 0.474 & 28.779 & 0.472\\
Src. dev. & 32.999 & 0.483 & 31.136 & 0.493 & 31.278 & 0.491\\
Adv. dev. & 50.666 & 0.578 & 48.752 & 0.600 & 48.816 & 0.598\\
Final & 51.259 & 0.592 & 49.572 & 0.611 & 49.574 & 0.610\\
\addlinespace
\multicolumn{7}{l}{\textbf{DeepLab (A2)}}\\*
Source & 21.614 & 0.387 & 20.323 & 0.391 & 20.198 & 0.390\\
Src. dev. & 20.908 & 0.387 & 19.753 & 0.390 & 19.578 & 0.389\\
Adv. dev. & 55.601 & 0.529 & 53.408 & 0.546 & 54.414 & 0.543\\
Final & 55.086 & 0.537 & 53.146 & 0.555 & 54.231 & 0.552\\
\addlinespace
\multicolumn{7}{l}{\textbf{OCRNet (A3)}}\\*
Source & 23.393 & 0.391 & 21.829 & 0.396 & 21.750 & 0.395\\
Src. dev. & 22.908 & 0.392 & 21.208 & 0.398 & 21.360 & 0.397\\
Adv. dev. & 46.993 & 0.508 & 44.143 & 0.524 & 44.990 & 0.523\\
Final & 47.008 & 0.504 & 44.093 & 0.522 & 44.794 & 0.518\\
\bottomrule\end{tabular}\end{center}
\caption{Nonempty-interior components at 90\% retention. Gains are reference minus MetaSeg: failure in percentage points; LL in $10^{-3}$ nats. Brackets show pointwise 95\% paired-image intervals. Comparators share eligible IDs and the recomputed budget within each row.}\label{tab:s-10}
\noindent\begin{minipage}{\linewidth}\footnotesize\raggedright
\noindent Source: \texttt{ELIGIBILITY\_MATCHED\_RESULTS.json}; 2,000 draws, seed 2026092501. Intervals use the separate eligibility-audit stream described in S4.
\end{minipage}
\end{table}

\clearpage
\pdfbookmark[0]{S5. Fixed-pair mapping and sensitivity analyses}{supp-s5}
\section*{S5. Fixed-pair mapping and sensitivity analyses}\label{sec:s5}
This analysis tests the Select question by changing the ranking rule without refitting the forecasts. It holds $(p,m)$ fixed and replaces native $m$ with $r_{\rm LL}(m,q)$. Both scores have access to the same available pair, but use it differently. The mapping is not generally monotone in $m$. All-valid evaluation remains primary; restriction to nonempty interiors is reported separately.

\begin{figure}[H]\centering
\includegraphics[width=390pt,trim=0 5.92bp 0 1.60bp,clip]{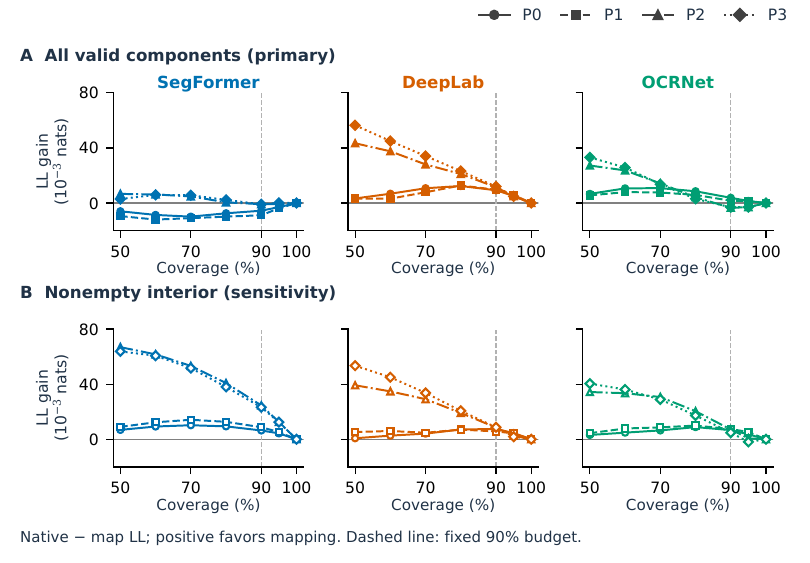}
\caption{Native-minus-mapped LL gains across coverage, in $10^{-3}$ nats: three architectures, four populations and two eligibility sets. All 168 archived estimates are shown; lines connect estimates. The dashed line marks primary 90\% coverage.}\label{fig:s-coverage}
\end{figure}

\noindent\textbf{Reading the results.} Tables S11--S12 compare rankings at 90\% retention. Table S13 and Figure~\ref{fig:s-coverage} vary the budget; Table S14 checks target-sequence dependence. All and Interior are evaluated with their own eligible sets and matched budgets.

\needroom\begin{table}[H]\centering
\begingroup\fontsize{8.5}{10.5}\selectfont\setlength{\tabcolsep}{3pt}\renewcommand{\arraystretch}{1.02}
\begin{tabular}{p{47.57pt}p{97.61pt}p{97.61pt}p{97.61pt}p{97.61pt}}
\toprule
Population & All: native-map LL & All: Entropy-map LL & Interior: native-map LL & All: native-map fail \\
\midrule
\multicolumn{5}{l}{\textbf{SegFormer (A1)}}\\*
Source & -5.547 \newline{}[-9.010,\allowbreak{} -2.389] & -13.952 \newline{}[-16.651,\allowbreak{} -11.387] & +6.420 \newline{}[+4.218,\allowbreak{} +8.601] & +0.643 \newline{}[+0.277,\allowbreak{} +0.997] \\
Src. dev. & -8.863 \newline{}[-12.422,\allowbreak{} -5.104] & -16.873 \newline{}[-20.011,\allowbreak{} -13.581] & +8.757 \newline{}[+6.083,\allowbreak{} +11.598] & +0.712 \newline{}[+0.326,\allowbreak{} +1.094] \\
Adv. dev. & -0.813 \newline{}[-3.323,\allowbreak{} +1.855] & -18.816 \newline{}[-21.148,\allowbreak{} -16.560] & +24.582 \newline{}[+22.424,\allowbreak{} +26.871] & -0.197 \newline{}[-0.453,\allowbreak{} +0.055] \\
Final & -0.926 \newline{}[-3.876,\allowbreak{} +2.083] & -17.209 \newline{}[-20.057,\allowbreak{} -14.374] & +23.236 \newline{}[+20.954,\allowbreak{} +25.544] & -0.126 \newline{}[-0.386,\allowbreak{} +0.125] \\
\multicolumn{5}{l}{\textbf{DeepLab (A2)}}\\*
Source & +9.281 \newline{}[+6.886,\allowbreak{} +11.669] & +3.412 \newline{}[+1.707,\allowbreak{} +5.193] & +7.495 \newline{}[+5.484,\allowbreak{} +9.553] & -1.499 \newline{}[-1.804,\allowbreak{} -1.211] \\
Src. dev. & +9.647 \newline{}[+6.196,\allowbreak{} +13.094] & +3.828 \newline{}[+1.211,\allowbreak{} +6.650] & +5.705 \newline{}[+2.887,\allowbreak{} +8.878] & -1.761 \newline{}[-2.118,\allowbreak{} -1.390] \\
Adv. dev. & +11.500 \newline{}[+9.561,\allowbreak{} +13.455] & -3.938 \newline{}[-5.199,\allowbreak{} -2.665] & +8.511 \newline{}[+6.457,\allowbreak{} +10.512] & -2.057 \newline{}[-2.287,\allowbreak{} -1.829] \\
Final & +11.969 \newline{}[+9.996,\allowbreak{} +14.036] & -4.096 \newline{}[-5.474,\allowbreak{} -2.650] & +8.707 \newline{}[+6.670,\allowbreak{} +10.568] & -2.361 \newline{}[-2.594,\allowbreak{} -2.136] \\
\multicolumn{5}{l}{\textbf{OCRNet (A3)}}\\*
Source & +3.811 \newline{}[+1.631,\allowbreak{} +6.069] & -3.140 \newline{}[-5.122,\allowbreak{} -0.943] & +6.696 \newline{}[+4.469,\allowbreak{} +8.932] & -1.009 \newline{}[-1.303,\allowbreak{} -0.716] \\
Src. dev. & +1.672 \newline{}[-1.689,\allowbreak{} +5.219] & -4.240 \newline{}[-6.978,\allowbreak{} -1.436] & +7.455 \newline{}[+4.282,\allowbreak{} +10.890] & -0.220 \newline{}[-0.567,\allowbreak{} +0.144] \\
Adv. dev. & -3.455 \newline{}[-5.698,\allowbreak{} -1.244] & -18.425 \newline{}[-20.197,\allowbreak{} -16.724] & +7.123 \newline{}[+4.931,\allowbreak{} +9.469] & -1.091 \newline{}[-1.317,\allowbreak{} -0.875] \\
Final & -3.068 \newline{}[-5.555,\allowbreak{} -0.824] & -17.216 \newline{}[-19.203,\allowbreak{} -15.318] & +4.834 \newline{}[+2.432,\allowbreak{} +7.123] & -1.299 \newline{}[-1.536,\allowbreak{} -1.056] \\
\bottomrule
\end{tabular}\endgroup
\caption{Mapping at 90\% retention: LL gains in $10^{-3}$ nats and failure gains in percentage points, with pointwise 95\% image intervals. Positive favors the mapping.}\label{tab:s-11}
\end{table}

\needroom\begin{table}[H]\centering
\begingroup\fontsize{8.5}{10.5}\selectfont\setlength{\tabcolsep}{3pt}\renewcommand{\arraystretch}{1.02}
\begin{tabular}{p{70pt}p{175pt}p{175pt}}
\toprule
Population & All-valid: $p$ minus mapped & Interior native minus map\\
\midrule
\multicolumn{3}{l}{\textbf{SegFormer (A1)}}\\*
Source & -0.504 [-0.716, -0.301] & -1.596 [-2.006, -1.178]\\
Src. dev. & -0.306 [-0.498, -0.107] & -1.409 [-1.856, -0.973]\\
Adv. dev. & -0.450 [-0.598, -0.297] & -1.566 [-1.849, -1.278]\\
Final & -0.232 [-0.382, -0.089] & -1.377 [-1.671, -1.082]\\
\multicolumn{3}{l}{\textbf{DeepLab (A2)}}\\*
Source & -1.623 [-1.899, -1.349] & -1.311 [-1.598, -1.014]\\
Src. dev. & -2.007 [-2.375, -1.634] & -1.301 [-1.691, -0.916]\\
Adv. dev. & -2.813 [-3.032, -2.587] & -2.024 [-2.235, -1.803]\\
Final & -3.008 [-3.235, -2.774] & -2.198 [-2.438, -1.953]\\
\multicolumn{3}{l}{\textbf{OCRNet (A3)}}\\*
Source & -1.356 [-1.618, -1.108] & -1.515 [-1.780, -1.246]\\
Src. dev. & -0.643 [-0.928, -0.368] & -1.330 [-1.679, -0.980]\\
Adv. dev. & -2.102 [-2.312, -1.893] & -2.822 [-3.083, -2.576]\\
Final & -2.286 [-2.537, -2.034] & -3.116 [-3.388, -2.850]\\
\bottomrule
\end{tabular}\endgroup
\caption{Failure comparisons completing Table S11. Entries are percentage points with pointwise 95\% intervals. The interior column uses its recomputed restricted budget.}\label{tab:s-12}
\end{table}

\needroom\begingroup\fontsize{8.5}{10.5}\selectfont\setlength{\tabcolsep}{3pt}\renewcommand{\arraystretch}{1.02}
\begin{longtable}{p{90pt}p{47.14pt}p{47.14pt}p{47.14pt}p{47.14pt}p{47.14pt}p{47.14pt}p{47.14pt}}
\toprule
Population / eligibility & 50\% & 60\% & 70\% & 80\% & 90\% & 95\% & 100\% \\
\midrule
\endfirsthead
\multicolumn{8}{l}{\footnotesize Table \thetable\ (continued)}\\
\toprule
Population / eligibility & 50\% & 60\% & 70\% & 80\% & 90\% & 95\% & 100\% \\
\midrule
\endhead

\multicolumn{8}{l}{\textbf{SegFormer (A1)}}\\*
Source /\allowbreak{} All & -5.962 & -8.611 & -9.861 & -7.419 & -5.547 & -2.931 & +0.000 \\*
Source /\allowbreak{} Interior & +6.837 & +9.285 & +10.218 & +9.392 & +6.420 & +4.299 & +0.000 \\*
Src. dev. /\allowbreak{} All & -9.458 & -11.981 & -10.937 & -9.857 & -8.863 & -3.773 & +0.000 \\*
Src. dev. /\allowbreak{} Interior & +9.006 & +12.366 & +14.174 & +12.687 & +8.757 & +5.282 & +0.000 \\*
Adv. dev. /\allowbreak{} All & +6.653 & +6.239 & +4.849 & +0.521 & -0.813 & -0.801 & +0.000 \\*
Adv. dev. /\allowbreak{} Interior & +66.861 & +61.632 & +53.254 & +40.817 & +24.582 & +12.594 & +0.000 \\*
Final /\allowbreak{} All & +2.941 & +5.941 & +5.658 & +2.304 & -0.926 & +0.149 & +0.000 \\*
Final /\allowbreak{} Interior & +63.779 & +60.646 & +51.690 & +38.026 & +23.236 & +12.431 & +0.000 \\
\multicolumn{8}{l}{\textbf{DeepLab (A2)}}\\*
Source /\allowbreak{} All & +3.339 & +6.758 & +10.755 & +12.383 & +9.281 & +5.163 & +0.000 \\*
Source /\allowbreak{} Interior & +0.764 & +2.670 & +4.132 & +7.140 & +7.495 & +4.582 & +0.000 \\*
Src. dev. /\allowbreak{} All & +3.249 & +3.188 & +7.852 & +12.763 & +9.647 & +5.913 & +0.000 \\*
Src. dev. /\allowbreak{} Interior & +5.278 & +6.049 & +4.742 & +7.060 & +5.705 & +4.354 & +0.000 \\*
Adv. dev. /\allowbreak{} All & +43.394 & +37.563 & +28.022 & +21.091 & +11.500 & +4.987 & +0.000 \\*
Adv. dev. /\allowbreak{} Interior & +39.195 & +34.779 & +29.125 & +19.140 & +8.511 & +2.565 & +0.000 \\*
Final /\allowbreak{} All & +56.452 & +45.003 & +34.120 & +23.209 & +11.969 & +4.916 & +0.000 \\*
Final /\allowbreak{} Interior & +53.415 & +45.030 & +33.648 & +20.672 & +8.707 & +1.876 & +0.000 \\
\multicolumn{8}{l}{\textbf{OCRNet (A3)}}\\*
Source /\allowbreak{} All & +6.669 & +10.659 & +10.825 & +8.566 & +3.811 & +1.508 & +0.000 \\*
Source /\allowbreak{} Interior & +3.151 & +4.840 & +6.456 & +8.859 & +6.696 & +3.603 & +0.000 \\*
Src. dev. /\allowbreak{} All & +5.565 & +7.961 & +7.651 & +5.919 & +1.672 & +1.445 & +0.000 \\*
Src. dev. /\allowbreak{} Interior & +4.459 & +7.844 & +8.660 & +9.969 & +7.455 & +5.284 & +0.000 \\*
Adv. dev. /\allowbreak{} All & +27.245 & +23.684 & +14.262 & +4.227 & -3.455 & -2.970 & +0.000 \\*
Adv. dev. /\allowbreak{} Interior & +34.305 & +33.465 & +30.683 & +20.408 & +7.123 & +0.696 & +0.000 \\*
Final /\allowbreak{} All & +33.151 & +25.776 & +13.650 & +3.071 & -3.068 & -2.896 & +0.000 \\*
Final /\allowbreak{} Interior & +40.436 & +36.124 & +28.942 & +17.284 & +4.834 & -1.895 & +0.000 \\
\bottomrule

\caption{Native-minus-mapped LL gains across coverage, in $10^{-3}$ nats. Entries are archived point estimates.}\label{tab:s-13}\\
\end{longtable}\endgroup

Table S14 resamples whole target sequences: P2 has 21 groups of 2--56 images and P3 has 33 groups of 1--43 images. Sampled image sums are divided by sampled evaluable-image counts, preserving the image-weighted estimand.

\needroom\begin{table}[H]\centering
\begingroup\fontsize{8.5}{10.5}\selectfont\setlength{\tabcolsep}{3pt}\renewcommand{\arraystretch}{1.02}
\begin{tabular}{p{111.0pt}p{111.0pt}p{111.0pt}p{111.0pt}}
\toprule
Population / eligibility & native-map LL & Entropy-map LL & native-map failure \\
\midrule
\multicolumn{4}{l}{\textbf{SegFormer (A1)}}\\*
Adv. dev. /\allowbreak{} All & -0.813 \newline{}[-3.759,\allowbreak{} +1.597] & -18.816 \newline{}[-21.481,\allowbreak{} -16.916] & -0.197 \newline{}[-0.421,\allowbreak{} +0.076] \\
Adv. dev. /\allowbreak{} Interior & +24.582 \newline{}[+21.474,\allowbreak{} +27.545] & +5.363 \newline{}[+3.087,\allowbreak{} +7.252] & -1.566 \newline{}[-1.749,\allowbreak{} -1.314] \\
Final /\allowbreak{} All & -0.926 \newline{}[-3.916,\allowbreak{} +2.537] & -17.209 \newline{}[-19.616,\allowbreak{} -14.402] & -0.126 \newline{}[-0.412,\allowbreak{} +0.135] \\
Final /\allowbreak{} Interior & +23.236 \newline{}[+21.019,\allowbreak{} +25.543] & +5.288 \newline{}[+3.925,\allowbreak{} +6.676] & -1.377 \newline{}[-1.682,\allowbreak{} -1.087] \\
\multicolumn{4}{l}{\textbf{DeepLab (A2)}}\\*
Adv. dev. /\allowbreak{} All & +11.500 \newline{}[+9.544,\allowbreak{} +14.077] & -3.938 \newline{}[-5.379,\allowbreak{} -2.281] & -2.057 \newline{}[-2.272,\allowbreak{} -1.852] \\
Adv. dev. /\allowbreak{} Interior & +8.511 \newline{}[+6.712,\allowbreak{} +10.521] & -5.155 \newline{}[-6.837,\allowbreak{} -3.316] & -2.024 \newline{}[-2.234,\allowbreak{} -1.763] \\
Final /\allowbreak{} All & +11.969 \newline{}[+9.679,\allowbreak{} +14.126] & -4.096 \newline{}[-6.179,\allowbreak{} -1.946] & -2.361 \newline{}[-2.564,\allowbreak{} -2.134] \\
Final /\allowbreak{} Interior & +8.707 \newline{}[+6.410,\allowbreak{} +11.010] & -5.583 \newline{}[-8.176,\allowbreak{} -3.010] & -2.198 \newline{}[-2.474,\allowbreak{} -1.938] \\
\multicolumn{4}{l}{\textbf{OCRNet (A3)}}\\*
Adv. dev. /\allowbreak{} All & -3.455 \newline{}[-7.111,\allowbreak{} +0.702] & -18.425 \newline{}[-22.478,\allowbreak{} -14.351] & -1.091 \newline{}[-1.340,\allowbreak{} -0.872] \\
Adv. dev. /\allowbreak{} Interior & +7.123 \newline{}[+2.311,\allowbreak{} +12.203] & -7.815 \newline{}[-12.724,\allowbreak{} -2.852] & -2.822 \newline{}[-3.099,\allowbreak{} -2.535] \\
Final /\allowbreak{} All & -3.068 \newline{}[-5.557,\allowbreak{} -0.412] & -17.216 \newline{}[-19.548,\allowbreak{} -14.678] & -1.299 \newline{}[-1.574,\allowbreak{} -0.995] \\
Final /\allowbreak{} Interior & +4.834 \newline{}[+0.982,\allowbreak{} +8.640] & -9.525 \newline{}[-13.225,\allowbreak{} -6.109] & -3.116 \newline{}[-3.378,\allowbreak{} -2.800] \\
\bottomrule
\end{tabular}\endgroup
\caption{Target mapping comparisons with whole-sequence 95\% intervals: 21 groups in P2 and 33 in P3. LL is in $10^{-3}$ nats; failure is in percentage points.}\label{tab:s-14}
\end{table}

\subsection*{Retained-set exchange accounting}
\begin{figure}[H]\centering
\includegraphics[width=.65\linewidth]{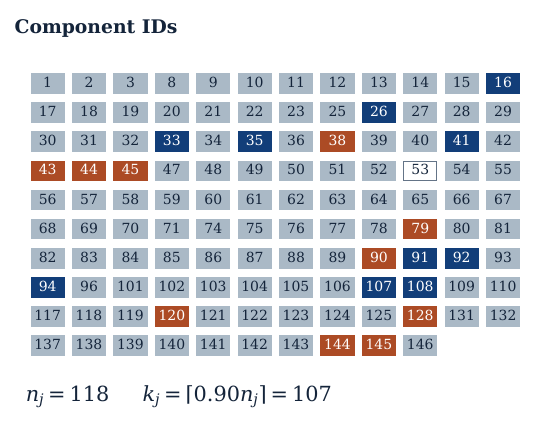}
\caption{Retained membership for the DeepLab source-development image in main Figure 7a. Each selector retains 107 of 118 components: 97 shared (gray), ten native-only (orange), ten mapped-only (blue); component 53 is rejected by both (open). Integer labels are original full-image identities.}\label{fig:s-case-ids}
\end{figure}

For each evaluable image, let $A_j=S_j(s_0)\setminus S_j(s)$ and $B_j=S_j(s)\setminus S_j(s_0)$. Their cardinalities are equal. Set $v_j=|A_j|/k_j$ and let $d_{\ell,j}$ be the difference between the mean losses on $A_j$ and $B_j$. Then
\[
\Delta_{\ell,j}=\frac{\sum_{i\in A_j}\ell_{ji}-\sum_{i\in B_j}\ell_{ji}}{k_j}=v_jd_{\ell,j},\qquad
\Delta_\ell=|J_E|^{-1}\sum_{j\in J_E}v_jd_{\ell,j}.
\]
When there is no exchange, $v_j=d_{\ell,j}=0$. The population gain is the mean of products, not the product of the displayed means. The identity attributes the loss difference exactly to changed membership. Figure~\ref{fig:s-case-ids} is a membership lookup for the Source 1 case, while Table~\ref{tab:s-loss-ledger} supplies the exclusive losses. The spatial interpretation is in main Figure 7.

\needroom\begin{table}[H]\centering
\begingroup\fontsize{8.5}{10.5}\selectfont\setlength{\tabcolsep}{3pt}\renewcommand{\arraystretch}{1.02}
\begin{tabular}{p{150.0pt}p{150.0pt}p{150.0pt}}
\toprule
Primary all-valid quantity & Minimum & Maximum \\
\midrule
Changed-image fraction (\%) & +98.500 & +100.000 \\
Mean exchanged retained fraction (\%) & +9.356 & +10.348 \\
\bottomrule
\end{tabular}\endgroup
\caption{Primary exchange ranges across the twelve architecture--population cells. These ranges summarize the cells and are not confidence intervals.}\label{tab:s-15}
\end{table}

\begin{figure}[H]\centering
\includegraphics[width=390pt,trim=0 0.88bp 0 2.32bp,clip]{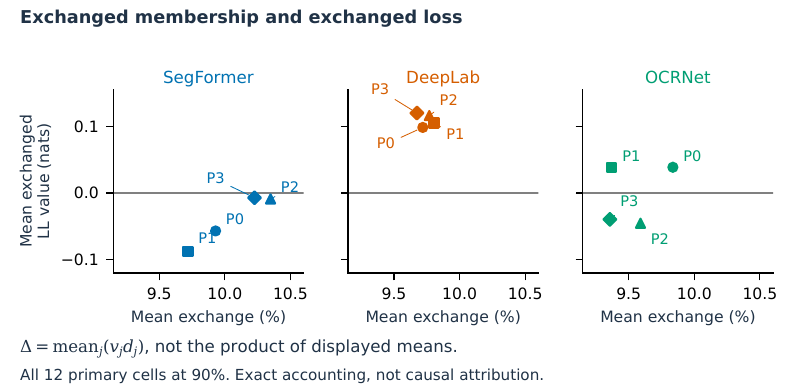}
\caption{Exchange descriptors for twelve primary comparisons at 90\% retention. P0/P1/P2/P3: source/source-development/adverse-development/final. Axes show mean exchanged fraction $v_j$ (\% of retained components) and mean exchanged LL difference $d_j$ (nats). Overall gain averages within-image products $v_jd_j$, not the product of these plotted means.}\label{fig:s-exchange}
\end{figure}

\Needspace{150pt}
\subsection*{Archived illustration audit}
Main Figure 7 uses six previously fixed Figure 2 identities and their original display crops: Source 1, Source 2, and the first archived fog, night, rain and snow examples. Identities were selected before reading case outcomes; their retained levels are in main Table 2. Insets magnify the smallest positive visible exclusive support in each crop, breaking ties by component ID. Inset choice uses geometry alone; masks are scaled separately for visibility, not area comparison. Selection uses the full eligible image at 90\% retention.

Main Figure 8 expands the twenty exclusive records in the Source 1 image of Figure 7a. Table~\ref{tab:s-loss-ledger} lists their IDs, binary outcomes and fixed-forecast losses. Figure~\ref{fig:s-case-ids} additionally shows the shared and unselected components in this image. The two exclusive groups are unpaired; each is a membership difference between the selectors, not an independently sampled population.

\needroom\begin{table}[H]\centering\begingroup\fontsize{9}{11}\selectfont\setlength{\tabcolsep}{8pt}
\begin{tabular}{@{}rrr@{}}
\toprule Component ID & $Y$ & $\ell_{\rm LL}(Y,p)$ [nats]\\\midrule
\multicolumn{3}{l}{\textit{Native-only}}\\*
38 & 0 & 0.414215\\*
43 & 0 & 0.282678\\*
44 & 0 & 0.435519\\*
45 & 0 & 0.628635\\*
79 & 0 & 0.399822\\*
90 & 0 & 0.427571\\*
120 & 0 & 0.479175\\*
128 & 0 & 1.082173\\*
144 & 1 & 0.290867\\*
145 & 1 & 0.460263\\
\addlinespace[3pt]
\multicolumn{3}{l}{\textit{Mapped-only}}\\*
16 & 1 & 0.205976\\*
26 & 0 & 0.984602\\*
33 & 1 & 0.383414\\*
35 & 0 & 1.072990\\*
41 & 1 & 0.413554\\*
91 & 1 & 0.296998\\*
92 & 1 & 0.386217\\*
94 & 1 & 0.434086\\*
107 & 0 & 1.218528\\*
108 & 0 & 1.221660\\
\addlinespace[3pt]
\bottomrule
\end{tabular}\endgroup
\caption{Twenty exclusive components from main Figure 8 (DeepLab/P1, Source 1). The native-only and mapped-only groups are unpaired and ordered by component ID. LL evaluates fixed $p$ (nats), rounded to six decimals; the accompanying numeric record preserves original precision. Values describe this single image.}\label{tab:s-loss-ledger}
\end{table}

\pdfbookmark[0]{S6. Original frozen-score evaluation}{supp-s6}
\section*{S6. Original frozen-score evaluation}\label{sec:s6}
The entropy reference is the source-normalized entropy of $p$ (R0). F1 uses source-normalized mean pixel entropy, F2 uses direct failure probability, and F3 adds the source-normalized calibration-opportunity probe with coefficient 0.25 to R0. The same frozen normalizers and full-source operators are applied to subsequent populations. P1 and P2 are development populations. The original P3 identities, operators and LL comparison matrix were fixed before the original outcome reveal. Later analyses on P3 are retrospective.

\needroom\begin{table}[H]\centering
\begingroup\fontsize{8.5}{10.5}\selectfont\setlength{\tabcolsep}{3pt}\renewcommand{\arraystretch}{1.02}
\begin{tabular}{p{111.0pt}p{111.0pt}p{111.0pt}p{111.0pt}}
\toprule
Population & F1 & F2=\allowbreak{}p & F3 \\
\midrule
\multicolumn{4}{l}{\textbf{SegFormer (A1)}}\\*
Src. dev. & -10.929 \newline{}[-13.887,\allowbreak{} -7.947] & -19.693 \newline{}[-23.135,\allowbreak{} -16.199] & -0.167 \newline{}[-0.912,\allowbreak{} +0.568] \\
Adv. dev. & -19.496 \newline{}[-21.347,\allowbreak{} -17.682] & -24.761 \newline{}[-27.233,\allowbreak{} -22.264] & -0.168 \newline{}[-0.648,\allowbreak{} +0.315] \\
Final & -18.026 \newline{}[-20.176,\allowbreak{} -15.839] & -20.890 \newline{}[-23.930,\allowbreak{} -17.662] & -0.911 \newline{}[-1.369,\allowbreak{} -0.416] \\
\multicolumn{4}{l}{\textbf{DeepLab (A2)}}\\*
Src. dev. & -6.826 \newline{}[-8.958,\allowbreak{} -4.561] & -7.703 \newline{}[-9.963,\allowbreak{} -5.242] & +0.194 \newline{}[-0.341,\allowbreak{} +0.756] \\
Adv. dev. & -22.036 \newline{}[-23.739,\allowbreak{} -20.368] & -16.711 \newline{}[-18.984,\allowbreak{} -14.511] & -1.966 \newline{}[-2.492,\allowbreak{} -1.423] \\
Final & -20.969 \newline{}[-22.725,\allowbreak{} -19.154] & -17.478 \newline{}[-19.799,\allowbreak{} -15.258] & -2.081 \newline{}[-2.578,\allowbreak{} -1.562] \\
\multicolumn{4}{l}{\textbf{OCRNet (A3)}}\\*
Src. dev. & -8.898 \newline{}[-11.184,\allowbreak{} -6.602] & -8.939 \newline{}[-11.715,\allowbreak{} -6.390] & -0.460 \newline{}[-1.219,\allowbreak{} +0.306] \\
Adv. dev. & -23.688 \newline{}[-25.504,\allowbreak{} -21.742] & -20.129 \newline{}[-22.940,\allowbreak{} -17.285] & -1.723 \newline{}[-2.464,\allowbreak{} -0.952] \\
Final & -20.280 \newline{}[-22.636,\allowbreak{} -17.881] & -21.554 \newline{}[-24.420,\allowbreak{} -18.651] & -2.915 \newline{}[-3.680,\allowbreak{} -2.127] \\
\bottomrule
\end{tabular}\endgroup
\caption{Complete original frozen-score LL@90 gains, R0 minus candidate, in $10^{-3}$ nats. Brackets are the original pointwise 95\% paired-image intervals. Source OOF is not pooled with deployment evaluation.}\label{tab:s-16}
\end{table}

\pdfbookmark[0]{S7. Additional development analyses}{supp-s7}
\section*{S7. Additional development analyses}\label{sec:s7}
This section reports additional analyses performed while developing the present research. They are part of this study, not a separate manuscript or prior publication. These analyses used two systems, 180 source-fit images, sixty development images and 429 later Cityscapes validation images. Deployment averaged five raw fold-derived operators, whereas the main comparisons use one full-source deployment fit. SegFormer normalization was image-balanced and DeepLab normalization pooled. The nominal class columns became constant after imputation, unlike the class indicators in the main comparisons. Source complementary-fit exposure also differed between systems. These results therefore describe the specified implementations, not an isolated architecture effect.

The additional analyses used ascending-score/ID retention, entropy of the observed probabilities in $(0,1)$, and their specified loss conventions; the DeepLab $10^{-7}$ guard was inactive on that range. Primary intervals were 90\%; subsequent direct-change analyses used 95\% intervals. The reported levels are preserved. These analyses document shared data and differences in fitting construction and are not counted as independent replication.

\begin{table}[H]\centering
\begingroup\fontsize{8.5}{10.5}\selectfont\setlength{\tabcolsep}{3pt}\renewcommand{\arraystretch}{1.02}
\begin{tabular}{p{150.0pt}p{150.0pt}p{150.0pt}}
\toprule
Cell & Gain x1000 nats \newline{}[90\% CI] & Positive single-fold operators \\
\midrule
Seg\_\allowbreak{}dev & -1.340 \newline{}[-2.706,\allowbreak{} +0.061] & 0/\allowbreak{}5 \\
Deep\_\allowbreak{}dev & +1.467 \newline{}[+0.313,\allowbreak{} +2.692] & 5/\allowbreak{}5 \\
Seg\_\allowbreak{}pros & -0.213 \newline{}[-0.597,\allowbreak{} +0.196] & 1/\allowbreak{}5 \\
Deep\_\allowbreak{}pros & -0.245 \newline{}[-0.715,\allowbreak{} +0.234] & 1/\allowbreak{}5 \\
\bottomrule
\end{tabular}\endgroup
\caption{Additional development-analysis gains with primary 90\% intervals. Favorable fold counts describe the fitted fold operators, not independent replications.}\label{tab:s-17}
\end{table}

\needroom\begin{table}[H]\centering
\begingroup\fontsize{8.5}{10.5}\selectfont\setlength{\tabcolsep}{3pt}\renewcommand{\arraystretch}{1.02}
\begin{tabular}{p{170pt}p{260pt}}
\toprule
Contrast & Point [interval]\\
\midrule
Image-weighted / T\_Seg & 1.126 [-0.446, 2.898]\\
Image-weighted / T\_Deep & -1.712 [-3.248, -0.202]\\
Image-weighted / Cdev & 2.807 [0.616, 5.123]\\
Image-weighted / Clater & -0.031 [-0.771, 0.738]\\
Image-weighted / D & -2.838 [-5.269, -0.500]\\
Equal-city / Seg\_dev & -1.340 [-2.996, 0.137]\\
Equal-city / Deep\_dev & 1.467 [0.121, 2.801]\\
Equal-city / Seg\_later & -0.483 [-1.239, 0.227]\\
Equal-city / Deep\_later & -0.093 [-1.089, 0.897]\\
Equal-city / Cdev & 2.807 [0.713, 5.018]\\
Equal-city / Clater & 0.390 [-0.651, 1.483]\\
Equal-city / D & -2.417 [-4.781, -0.113]\\
Equal-city / T\_Seg & 0.856 [-0.735, 2.601]\\
Equal-city / T\_Deep & -1.560 [-3.266, 0.105]\\
Direct-LL / Seg\_dev / probe & -2.716 [-5.043, -0.232]\\
Direct-LL / Seg\_dev / fusion & -0.998 [-2.140, -0.018]\\
Direct-LL / Deep\_dev / probe & -3.320 [-5.678, -0.952]\\
Direct-LL / Deep\_dev / fusion & 0.153 [-0.456, 0.750]\\
Direct-LL / Seg\_later / probe & -2.447 [-3.224, -1.674]\\
Direct-LL / Seg\_later / fusion & -0.412 [-0.648, -0.194]\\
Direct-LL / Deep\_later / probe & -3.421 [-4.264, -2.623]\\
Direct-LL / Deep\_later / fusion & -0.319 [-0.606, -0.031]\\
\bottomrule
\end{tabular}\endgroup
\caption{Additional direct-change and training-target controls, in $10^{-3}$ nats. $T$ is later minus development gain; $C$ is DeepLab minus SegFormer gain; $D=C_{\rm later}-C_{\rm dev}$. Direct-change and equal-city intervals are 95\%; direct-LL controls retain their original 90\% intervals. Equal-city weighting is a sensitivity estimand.}\label{tab:s-18}
\end{table}

\pdfbookmark[0]{S8. Evidence chronology and reproducibility}{supp-s8}
\section*{S8. Evidence chronology and reproducibility}\label{sec:s8}
\subsection*{Analysis chronology}
\begin{specblock}{Evidence status by analysis.}
Original P0--P2 & Outcomes historically exposed.\\
Original P3 matrix & Identities, operators and LL comparisons frozen before its original outcome reveal.\\
Later analyses & MetaSeg, eligibility completion, fixed-pair mapping and later P3 diagnostics are retrospective.\\
Adaptation freeze & Source probabilities frozen before its own target-outcome read; project-level exposure remains as above.\\
\end{specblock}

\begin{specblock}{Evidence status of the objects listed in Table S3.}
Base forecast $p$ (F2) & original P3 sealed matrix only\\
Frozen score F3 & original frozen candidate\\
Descriptors (G3) & post-hoc constructive source control\\
All fields (G4) & post-hoc constructive source control\\
MetaSeg & adaptation; later P3 post-hoc\\
Fixed-pair mapping & post-hoc matched action audit\\
\end{specblock}

\paragraph{Uncertainty and dependence}
The 95\% intervals are nominal, pointwise and conditional on the recorded images and fitted operators. They omit repeated-training variation and multiplicity correction. An interval crossing zero leaves the sign unresolved; it does not establish equivalence. S5 supplies sequence resampling for mapping on P2/P3. Image-name separation alone does not establish scene or sequence independence. An earlier inventory found no group-clean reserve in the inspected local ACDC remainder; this finding does not extend to other data resources.

\paragraph{Corrected source comparison}
Strict source results use highest-entropy rejection and fully nested upstream inputs. An earlier non-strict analysis used an incorrectly oriented entropy comparator and different input dependencies; its external-transport values are excluded from strict-source evidence. The corrected source entropy-reference LL@90 values are 0.480096, 0.418753 and 0.422593 nats for SegFormer, DeepLab and OCRNet, respectively (rounded for display; full precision is retained in the archived results). Input-freeze digests were checked across all 75 upstream fits and 240 downstream designs. Historical corrections remain in the package provenance.

\subsection*{Reproducibility and availability}
\begin{specblock}{Manuscript sources and the public research archive.}
Manuscript source package & LaTeX documents and figure assets for the main paper and supplement. Compilation requires no model training, inference or statistical estimation.\\
Public research archive & Evaluation software, synthetic tests, experimental specifications and numerical evidence at \url{https://doi.org/10.5281/zenodo.23276040}. The archive README and \path{docs/RESULT_INDEX.md} identify the materials.\\
Portable evaluator & \path{evaluator/evaluate.py} accepts aligned IDs, probabilities, scores, masks and recorded draws. Synthetic checks include the integer-coverage fixture; \path{docs/REPLAY_SCOPE.md} describes the supported scope.\\
Evaluation checks & Probability bounds; finite scores; binary eligible outcomes; duplicate IDs; clipping; score direction and ties; integer coverage; empty images; image-weighted sequence resampling.\\
Returned quantities & Retained identities, endpoint losses and exact exchange accounting under the supplied resampling design.\\
Real-cache check & A1/P0, all-valid native-versus-map comparison at 90\%; retained-ID and interval agreement. This is distinct from the synthetic tests.\\
\end{specblock}

Raw images, labels, checkpoints and restricted component records are not redistributed. Upstream fitting and the complete research pipeline have not been independently rerun. Software versions underlying some implicit library defaults, and the additional analyses' source-selection and mixing provenance, remain incompletely recovered; the current environment does not establish these details. The documentation distinguishes portable checks, authorized-cache replay and described upstream computations.

\FloatBarrier
\pdfbookmark[0]{S9. Literature positioning}{supp-s9}
\section*{S9. Literature positioning}
Table~\ref{tab:s-literature} positions the comparison designs within their predictive and decision-theoretic foundations, by evaluation unit and target. MetaSeg, source controls and fixed-pair mapping supply this study's experimental evidence; the other citations provide the surrounding context.

\begin{table}[H]\centering
\begingroup\fontsize{9}{11}\selectfont\setlength{\tabcolsep}{4pt}\renewcommand{\arraystretch}{1.12}
\begin{tabular}{@{}p{.24\linewidth}p{.34\linewidth}p{.37\linewidth}@{}}
\toprule Research line and unit & Established evaluation target & Connection to this study\\\midrule
Predictive uncertainty; samples or pixels~\cite{gal2016,kendall2017,lakshminarayanan2017,ovadia2019,dejorge2023} & Model and data uncertainty; calibration, error detection and distribution shift. & Frozen segmentation outputs; component losses and per-image retention budgets.\\[5pt]
Failure prediction; pixels, components or images~\cite{hendrycks2017,corbiere2019,rottmann2020,zenk2025,kahl2024} & Correctness or quality forecasts; confidence aggregation to the decision unit. & Component MetaSeg; entropy and probability references; prediction and selection assessed separately.\\[5pt]
Selective prediction; samples with rejection~\cite{elyaniv2010,hendrickx2024,geifman2017,geifman2019,franc2023,liang2024,cattelan2024} & Risk--coverage, conditional-risk selection, joint predictor/rejector learning and shift. & A fixed predictor and matched retained counts within each image.\\[5pt]
Probability calibration; class forecasts~\cite{guo2017,ding2021,wang2023,zhao2021,zhu2024} & Class-probability calibration, spatial scaling and decision-specific calibration. & Component-failure forecasts; diagnostic CE; retained LL of unchanged base $p$.\\[5pt]
Evaluation; accepted sets and thresholds~\cite{condessa2017,hanczar2019,ding2020,galil2023,traub2024} & Calibration, discrimination and selection measures; rejection; multi-threshold alignment. & Absolute retained levels, paired gains, coverage and eligibility sensitivity.\\[5pt]
Loss alignment; forecasts and image-level segmentation~\cite{gneiting2007,franc2023,borges2026} & Proper probability scores, conditional-risk ordering and Dice-aligned confidence. & Component failure versus forecast loss; native versus LL-risk ranking on fixed $(p,m)$.\\
\bottomrule
\end{tabular}\endgroup
\caption{Literature map by evaluation unit and target. The final column identifies the corresponding comparison in this study.}\label{tab:s-literature}
\end{table}